\documentclass[10pt,twocolumn,letterpaper]{article}
\usepackage[pagenumbers]{cvpr}
\usepackage{amsmath,amsfonts}
\usepackage{algorithmic}
\usepackage{algorithm}
\usepackage{array}
\usepackage{textcomp}
\usepackage{stfloats}
\usepackage{url}
\usepackage{ifthen}
\usepackage{verbatim}
\usepackage{graphicx}
\usepackage{xcolor}
\usepackage{cite}
\usepackage{booktabs}
\usepackage{makecell}
\usepackage{multirow}
\definecolor{forestgreen}{RGB}{50,139,34}
\newcommand{\thename}{ViTMatte}
\usepackage[pagebackref,breaklinks,colorlinks]{hyperref}

\begin{document}

\title{ViTMatte: Boosting Image Matting with Pretrained Plain Vision Transformers}

\author{
Jingfeng Yao$^{1,\star}$,
Xinggang Wang$^{1,\dagger}$,
Shusheng Yang$^{1}$,
Baoyuan Wang$^{2}$
\\
[2mm]
$^1$~School of EIC, Huazhong University of Science \& Technology\\ 
$^2$~Xiaobing.AI
\\ 
\normalsize{
\texttt{\{jingfengyao,xgwang@hust.edu.cn\}}}}

\maketitle

\let\thefootnote\relax\footnotetext{$^\star$This work was done when Jingfeng Yao was interning at Xiaobing.AI. $^\dagger$Xinggang Wang is the corresponding author: \texttt{xgwang@hust.edu.cn}}

\begin{abstract}
Recently, plain vision Transformers (ViTs) have shown impressive performance on various computer vision tasks, thanks to their strong modeling capacity and large-scale pretraining. However, they have not yet conquered the problem of image matting. We hypothesize that image matting could also be boosted by ViTs and present a new efficient and robust ViT-based matting system, named \thename{}. Our method utilizes (i) a hybrid attention mechanism combined with a convolution neck to help ViTs achieve an excellent performance-computation trade-off in matting tasks. (ii) Additionally, we introduce the detail capture module, which just consists of simple lightweight convolutions to complement the detailed information required by matting. To the best of our knowledge, \thename{} is the first work to unleash the potential of ViT on image matting with concise adaptation. It inherits many superior properties from ViT to matting, including various pretraining strategies, concise architecture design, and flexible inference strategies. We evaluate \thename{} on Composition-1k and Distinctions-646, the most commonly used benchmark for image matting, our method achieves state-of-the-art performance and outperforms prior matting works by a large margin. Our codes and models are available at: \href{https://github.com/hustvl/ViTMatte}{https://github.com/hustvl/ViTMatte}.
\end{abstract}

\section{Introduction}

\begin{figure}
    \centering
    \includegraphics[width=0.9\linewidth]{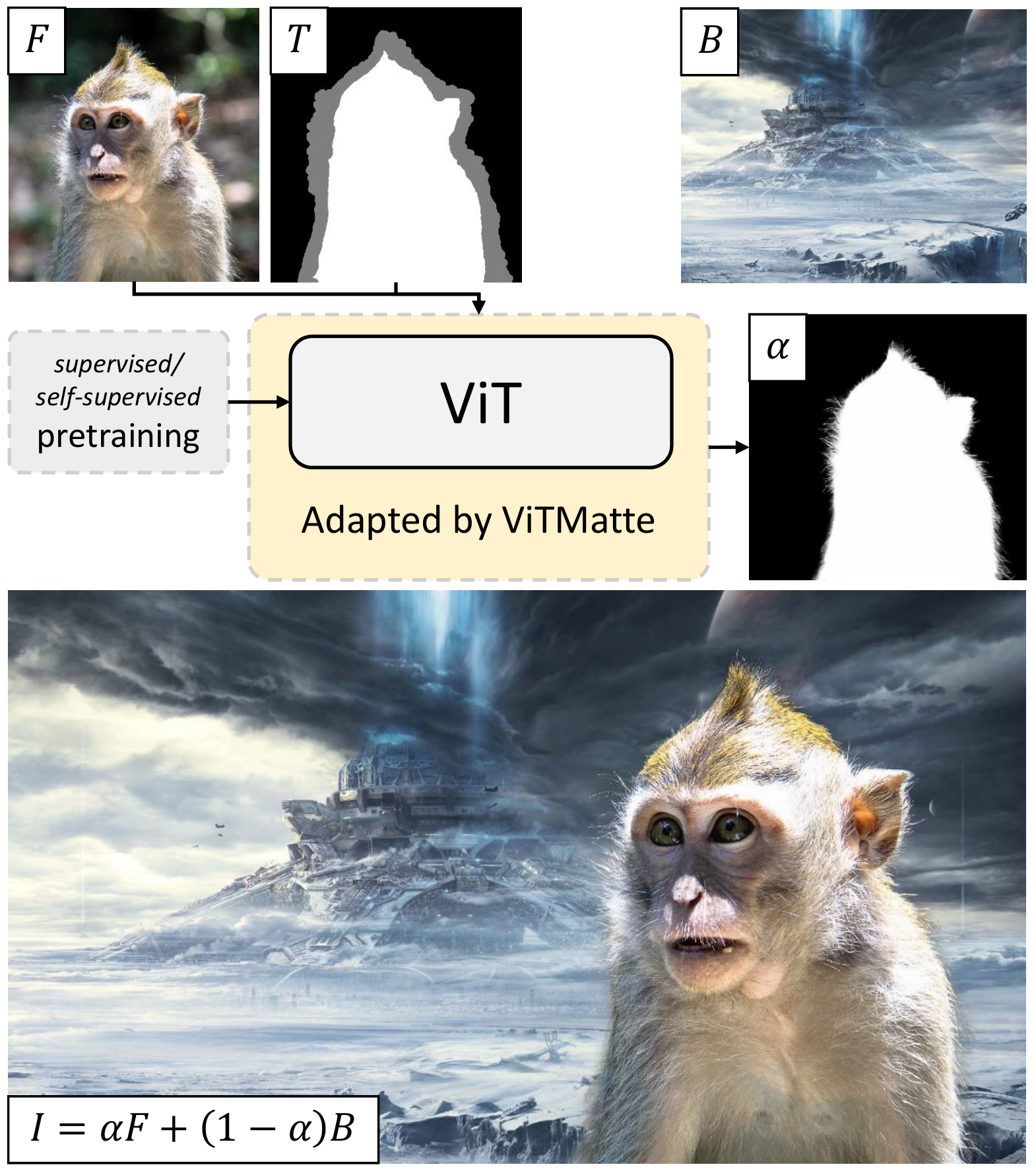}
    \caption{Pipeline of image matting. We input foreground $F$ and its corresponding trimap $T$ into \thename{} and predict the alpha matte $\alpha$. Then we can use them to create a new composition image with the equation $I=\alpha F+(1-\alpha)B$. The images are from AIM-500~\cite{aim500} and the internet. Please zoom in for a better view.}
    \label{fig:first}
\end{figure}

Image matting has been a long-standing and fundamental research problem in computer vision \cite{closed-form, matting_survey}. As shown in Figure~\ref{fig:first}, it aims to precisely separate the foreground object and background by predicting the alpha matte for each pixel (also known as Alpha Mating). It can be applied to numerous killer applications, such as movie special effects, digital person creation, video conferences, and so on. In recent years, the performance of image matting has been dramatically improved by deep learning-based methods \cite{DIM, sim, MGM, Tripartitle2021} which can leverage the strong semantic representation to capture meaningful context, compared with traditional sampling~\cite{matting2013, hematting2011} or propagation-based methods \cite{poissonmatting04, nonlocalmatting2011, knnmatting2013}. The mainstream CNN-based Image matting typically follows such a paradigm: A hierarchical backbone is used to extract the features of the image and then a decoder with injected prior is employed to fuse multi-level features. It is generally assumed that the decoder needs to simultaneously accomplish two tasks (i) fusing multi-level features and (ii) capturing the detailed information \cite{HAttMatting, GCAMatting,MGM}, which complicates the design of the decoder and whole system.

\begin{figure*}
    \centering
    \includegraphics[width=0.9\textwidth]{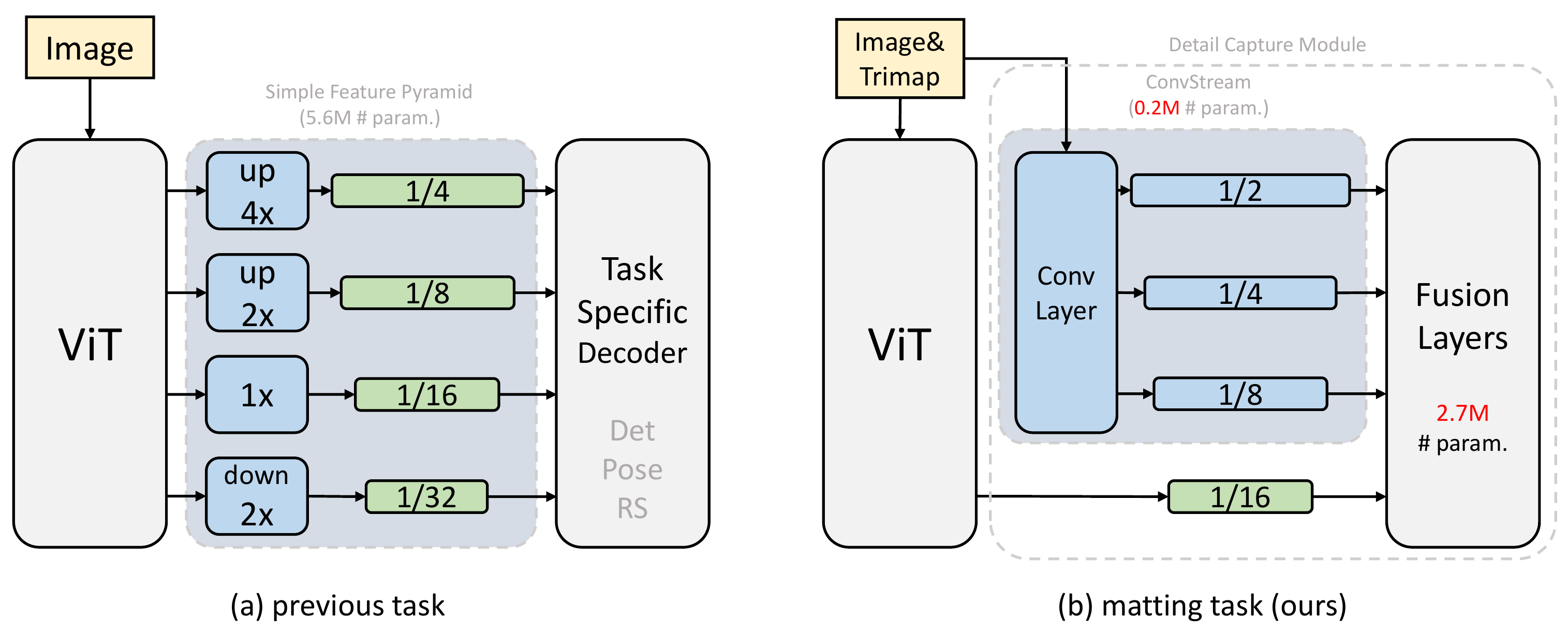}
    \caption{Overview of \thename{} and other applications of plain vision transformers~\cite{vitdet, vitpose, vitrs}. They use a simple feature pyramid designed by ViTDet~\cite{vitdet}. Differently, we propose a new adaptation strategy, especially for image matting, named \thename{}. We use simple convolution layers to get detailed information about the image and the feature map output by plain vision transformers is used only once.}
    \label{fig:vit_apps}
\end{figure*}

On the other hand, the plain vision transformer (ViT) has become a powerful backbone for various computer vision tasks~\cite{vit, vitdet, vitpose}. Unlike the commonly used hierarchical backbone~\cite{res, swin}, ViT is minimalist and \emph{non-hierarchical}. Recently, some works have explored the use of plain vision transformer on object detection~\cite{vitdet} and pose estimation~\cite{vitpose} and achieved remarkable results. The key insight behind this is that a task-agnostic pretrained transformer structure could already encode rich enough semantic representation which simplifies the adaptation of downstream tasks. For example, ViTDet \cite{vitdet} finds that even without the feature fusion process of a Feature Pyramid Network (FPN)~\cite{fpn}, ViT is still able to achieve impressive performance with a feature pyramid generated by simple deconvolutions. A similar paradigm shift is also observed in other domains, where the foundation models (i.e. GPT-3 ~\cite{GPT3} and Florence ~\cite{Florence}) are supposed to do most of the heavy lifting things. Inspired by all those prior works, it would be interesting to raise the question: \emph{if a plain ViT is ``foundation" enough for solving image matting with concise adaptation?}

In this paper, we try to enable the ViTs to be finetuned on matting tasks and unleash their potential. Our goal is not to design new complex modules tailored for image matting but to pursue more general and effective matting architectures with minimal adaptations. If successful, it would further confirm the paradigm-shifting and decouple the task-agnostic pre-training from task-specific adaptation. However, to explore ViT for image matting, there are two specific challenges. (1) How to reduce the intensive computational cost for high-resolution images ~\cite{HDMatt}? Plain ViT computes self-attention among all patches in the image, producing long patch sequences and excessive computational burden. (2) How to capture the finest details on top of the non-hierarchical ViT representation. Considering the motivation discussed above, we are \emph{not} supposed to carefully design complex hierarchical feature fusion mechanisms as in prior works ~\cite{GCAMatting, MGM, matteformer, sim}.

To address the challenges above, we propose \thename, an efficient and effective image matting system with a plain vision transformer. Based on our analysis, we believe that a pretrained ViT model provides the majority of the necessary functionality for image matting, and we only need to make concise and lightweight adaptations to use it for this purpose. On one hand, a plain ViT is stacked with the same transformer blocks, each of which computes the global self-attention at an expensive cost. We argue that it is unnecessary and propose a simple ViT adaptation strategy tailored for solving matting. Specifically, we employ both window and global attention to achieve better trade-offs for the computation. Besides, we find that the convolution module can effectively enhance the global attention on top of ViT, and the residual convolutional neck could further improve matting performance. On the other hand, a plain ViT has a fixed patch embedding process, leading to information loss, especially for very subtle details. To fully model the details for matting, we introduce a detail capture module specially for plain ViTs, which consists of only less than 3M number of parameters.

Compared with previous ViT-based tasks and matting systems, \thename{} is the first ViT-adaptation method specially designed for image matting. As shown in Figure~\ref{fig:vit_apps}, compared to the previous adaptation strategy~\cite{vitdet}, it gets better results with fewer parameters. It could save 70\% FLOPs when processing high-resolution images. \thename{} is also the first ViT-based image matting method and boosts image matting with various self-supervised pretrained ViTs. We evaluate \thename{} on the most widely-used benchmark Composition-1k and Distinctions-646, and it achieves the new state-of-the-art results with fewer parameters. Our contributions can be summarized as follows:

\begin{itemize}
    \item We propose \thename{}, the first plain ViT-based matting system. To address the challenges, we introduce a ViT adaptation strategy and a detail capture module. For the first time, we prove that a plain vision transformer can achieve significantly better matting performance than other backbones with even fewer parameters.
    \item We evaluate \thename{} on the most widely used benchmarks. Compared with previous state-of-the-art methods, our model gets 2.54 improvement of SAD and 3.06 improvement of Connectivity on Composition-1k; and gets 8.60 improvements of SAD and 8.50 improvement of Connectivity on Distinctions-646 respectively, making \thename{} as the new \emph{state-of-the-art} system with even smaller model size.
    \item \thename{} well inherits the advantages of ViTs. Plenty of comprehensive experiments and analyses have been done to compare it to previous ViT-adaptation strategies and previous matting systems. It reveals unique insights from ViTMatte. We hope it could inspire the following matting work with plain vision transformers.
\end{itemize}

\section{Related Work}
Here we mainly review the most relevant works, refer to ~\cite{matting_survey, rvm, matteformer} for a detailed discussion of the matting problem.

\subsection{Transformer-based Image Matting}
Learning-based image matting has been dominated by convolutional neural networks (CNN) in the literature \cite{DIM, CAM, HAttMatting, GCAMatting, MGM, Tang_2019_CVPR, Zhang_2019_CVPR, rvm, Lin_2022_WACV, Wang_2021_ICCV} for a long time. Until recently, transformer-based \cite{attention,vit} methods disrupted a wide spectrum of many vision tasks thanks to their powerful long-distance modeling capability compared with CNNs. Inspired by the paradigm-shifting, a few recent works started to employ transformers for solving matting tasks \cite{matteformer, rmat, transmatting} and showed encouraging results, i.e. Swin Transformer~\cite{swin} and SegFormer~\cite{segformer}. However, these specialized visual transformers have a similar hierarchical structure as CNNs and are designed as direct replacements for CNN backbones. Technologies are involved rapidly, the recent conclusion is drawn by \cite{vitdet} indicating that a plain ViT could be more powerful than expected despite its \emph{minimalist} and \emph{non-hierarchical} nature. \cite{vitdet} reveals an important message that a vision foundation model \cite{bommasani2021opportunities} can be trained based on plain ViT and the downstream tasks can be accomplished by task-specific adaptation. In this paper, we aim to study the difficult image matting task since it requires detailed visual information that is not easy to be learned in a foundation model.

\subsection{Pre-training of Plain ViTs}
Pre-training and fine-tuning have been the de facto paradigm for many visual understanding tasks. Most vision transformers are typically pre-trained on ImageNet~\cite{deng2009imagenet} with supervised learning. Recently, self-supervised pre-training strategies have been introduced to the vision from NLP~\cite{devlin2018bert, radford2019language, brown2020language, radford2018improving}. They address the appetite for data problems by using unlabeled data. Many of them, such as MAE~\cite{he2022masked}, DINO~\cite{dino}, and iBOT~\cite{ibot} mainly target plain vision transformer structures and pretrained models in a self-supervised manner. These methods have been shown to be able to facilitate many downstream tasks, such as semantic segmentation, object detection, and instance segmentation. However, how to best leverage such pretrained representation for matting is still yet to be explored to a large extent from both computational cost and accuracy perspectives.

\subsection{Plain ViTs for Downstream Tasks}
The plain vision transformer is originally used as a strong backbone for image classification\cite{vit}. However, due to its non-hierarchical structure design, it has poor compatibility with common decoders or heads for many downstream tasks. People prefer to use transformers that are specifically designed for vision tasks such as~\cite{swin, PVT, T2T,heo2021rethinking,fan2021multiscale}. Their multi-level architecture makes them directly transferable to many convolution-based tasks. However, with the rise of self-supervised pre-training for plain ViT(i.e.~\cite{he2022masked}), there has been a renewed focus on this non-hierarchical structure. For example, ViTDet\cite{vitdet} finds that using only parallel deconvolutions to generate the simple feature pyramid without the heavy feature pyramid networks (FPN)~\cite{fpn} allows plain ViT to achieve impressive results on object detection tasks.  ViTPose~\cite{vitpose} find that ViTs are more compatible with simple decoder designs than convolutional neural networks. We speculate that this concise property of ViT could facilitate a new structural design for solving image matting.

\section{Methodology} 

\subsection{Preliminary}
To improve clarity, we provide a concise explanation of the concepts of trimap in image matting and plain vision transformers.

\begin{figure}
    \centering
    \includegraphics[width=0.9\linewidth]{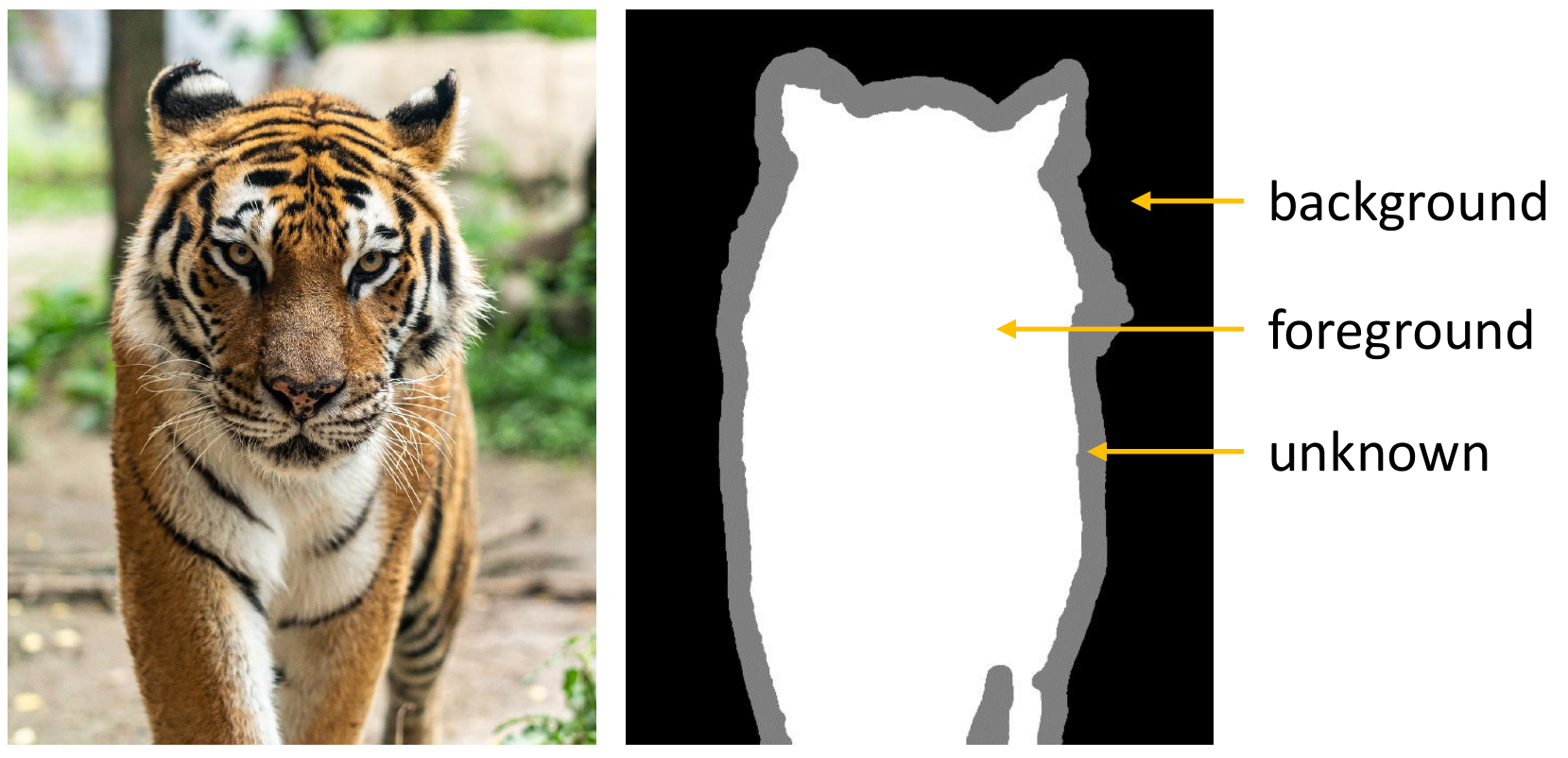}
    \caption{Original Image (left) and its corresponding trimap (right). The trimap is the most widely-used manually drawn hint map for image matting.}
    \label{fig:trimap}
\end{figure}

\textbf{Trimap in Image Matting} 
In natural image matting, researchers always use trimaps as priors to distinguish foreground and background. Trimap is a manually labeled hint map for image matting. As shown in Figure~\ref{fig:trimap}, users need to simply outline the foreground and background for a given image and the matting algorithm then calculates the transparency or so-called alpha values only in the rest unknown regions. It is a commonly used hint for natural image matting for a long time~\cite{knn, closed-form} and carried over in the deep learning era~\cite{matteformer, rmat, transmatting}. In \thename{}, we use trimaps in the form of grayscale images and it's noted as $T\in R^{H\times W\times 1}$.

\begin{equation}
    T_{(x,y)}=
    {\begin{cases}
    0&           \text{$(x,y) \in background$}\\
    1&           \text{$(x,y) \in foreground$}\\
    0.5&         \text{$(x,y) \in unknown$}
    \end{cases}}
\end{equation}

\textbf{Plain Vision Transformers}
Plain vision transformer here refers specifically to the architecture proposed by Dosovitskiy et al.~\cite{vit} and not to other variants designed for vision. It is a non-hierarchical architecture and only provides output with the same size as the input.

Given input image $x\in R^{H\times W\times C}$, Where $(H, W)$ denotes the input image resolution and $C$ denotes its number of channels, ViT flattens and embeds it into a sequence of image tokens $x_{p_0}\in R^{N\times(P^2\times C)}$ by a linear patch embedding layer. $P$ denotes patch size and $N=HW/P^2$ denotes the number of image tokens. The image sequence is fed into vision transformers. A transformer layer consists of multi-head self-attention (MHSA)~\cite{NIPS2017_3f5ee243} and MLP blocks. The residual connections and LayerNorm (LN) are applied to it. Equation~\eqref {eq:transformer1} and~\eqref{eq:transformer2} illustrate one layer of vision transformer.

\begin{equation}
    x_{p_l}' = MHSA(LN(x_{p_l})) + x_{p_l}\\
    \label{eq:transformer1}
\end{equation}
\begin{equation}
    \label{eq:transformer2}
x_{p_{l+1}} = MLP(LN(x_{p_l}')) + x_{p_l}'
\end{equation}

A plain vision transformer generates a sequence of image tokens ${x_{p_1}, ..., x_{p_L}}$, where $L$ represents the number of layers in the vision transformer, and all tokens are of the same size as the input $x_{p_0}$. Typically, the final feature $x_{p_L}$ is used as the output of the plain vision transformer.

\subsection{Overall Architecture}
Figure~\ref{fig:vit_apps} illustrates our proposed \thename{}, a concise and efficient image matting system based on plain vision transformers. Given an RGB image $X\in R^{H\times W\times 3}$ and its corresponding trimap $T\in R^{H\times W\times 1}$, we concatenate them channel-wise and input them to \thename{}. Our system extracts multi-level features using plain vision transformers and a detail capture module. The plain vision transformer serves as the foundational feature extractor and generates only a single feature map with a stride of 16. The detail capture module consists of a cascade of convolutional layers that capture and fuse detailed information for image matting. We eschew specialized designs~\cite{MGM} and instead, simply upsample and fuse features at different scales to predict the final alpha matte $\alpha \in R^{H\times W\times 1}$.

\subsection{Vision Transformer Adaptation}

\begin{figure}[tbp]
    \centering
    \includegraphics[width=0.9\linewidth]{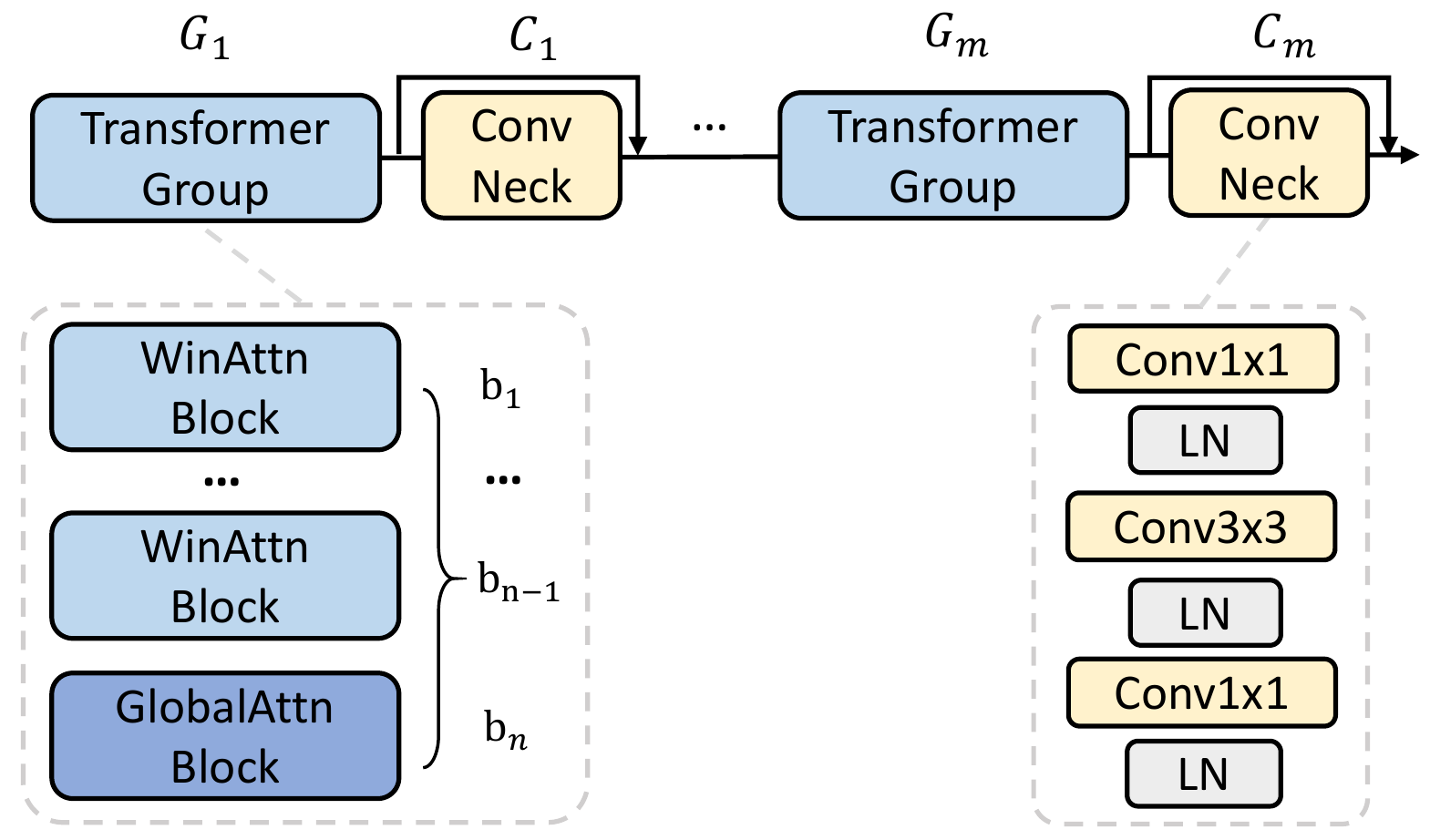}
    \caption{Backbone adaptation of \thename{}. We evenly use window attention and global attention in vision transformer layers to reduce computation burden and add convolution necks to enhance more detail information for matting.}
    \label{fig:backbone_adaptation}
\end{figure}

\thename{} enhances the vision transformer for image matting by using the hybrid attention mechanism and adding a convolution neck between transformer blocks. Figure~\ref{fig:backbone_adaptation} illustrate our adaptation strategy.

We suppose that computing global self-attention is unnecessary for image matting, as it can introduce prohibitively high computational complexity for high-resolution images, as shown in Equation~\eqref{comp_self_attn}. Inspired by Li et al.~\cite{vitdet}, we propose a hybrid attention mechanism for plain vision transformers.

Specifically, we divide the blocks in the plain ViT into $m$ groups $G=\{G_1, G_2, ..., G_m \}$. Each group contains $n$ transformer blocks denoted as $G_i=\{b_1, b_2, ..., b_n\}$, while $G_i\in G$. For blocks in $G_i$, we only apply global attention in the last block $b_n$, while using window attention instead of global attention in the other blocks $\{b_1, b_2, ..., b_{n-1}\}$ The window size is denoted as $k$, and the computational complexity of this mechanism is shown in Equation~\eqref{comp_win_attn}.

\begin{equation}
    O((HW)\cdot(HW)\cdot C)
    \label{comp_self_attn}
\end{equation}
\begin{equation}
     O(k^2 \cdot k^2 \cdot C)
    \label{comp_win_attn}
\end{equation}

By adapting the transformer blocks in this manner, we can greatly reduce the computational complexity of ViT, particularly for high-resolution images.

Furthermore, we incorporate a convolutional block after each group of transformer blocks, denoted as $C=\{C_1, C_2, ..., C_m \}$, and utilize a residual connection to feed forward the results of each group. The number of convolution blocks is equal to the group number. To achieve this, we employ the ResBottleNeck \cite{res} block as our convolutional block design, as it has been demonstrated to be effective for matting tasks~\cite{GCAMatting, MGM}.

We hypothesize the reason is that attention mechanisms tend to pay more attention to low-frequency information \cite{iformer}. The convolutional block there could serve as a high-pass filter to extract high-frequency information such as boundaries or textures, which are crucial for accurate matting. By utilizing a residual connection, we are able to preserve the low-frequency information captured by the transformer blocks, while enhancing the high-frequency information through the convolutional block.

\subsection{Detail Capture Module}
\label{Detail Capture Module}

Given that the vision transformer performs the majority of the required tasks, we have incorporated a lightweight detail capture module to effectively capture finer details. This module comprises a convolution stream and a straightforward fusion strategy, which work together to supplement the missing detailed information.

In Figure~\ref{fig:vit_apps}, we demonstrate the use of a convolution stream to capture detailed information in the image. The convolution stream is comprised of a sequence of simple \textit{conv3$\times$3} layers, which are designed to capture finer details. Each layer includes a convolutional layer with a kernel size of 3, along with batch normalization and a ReLU activation function. The length of the ConvStream is set at 3. For an input $\in R^{H\times W\times C}$, the convstem produces a set of detailed feature maps $D=\{D_1, D_2, D_3\}$ at resolutions of $\{\frac{1}{2}, \frac{1}{4}, \frac{1}{8} \}$.

\begin{figure}[tbp]
    \centering
    \includegraphics[width=0.7\linewidth]{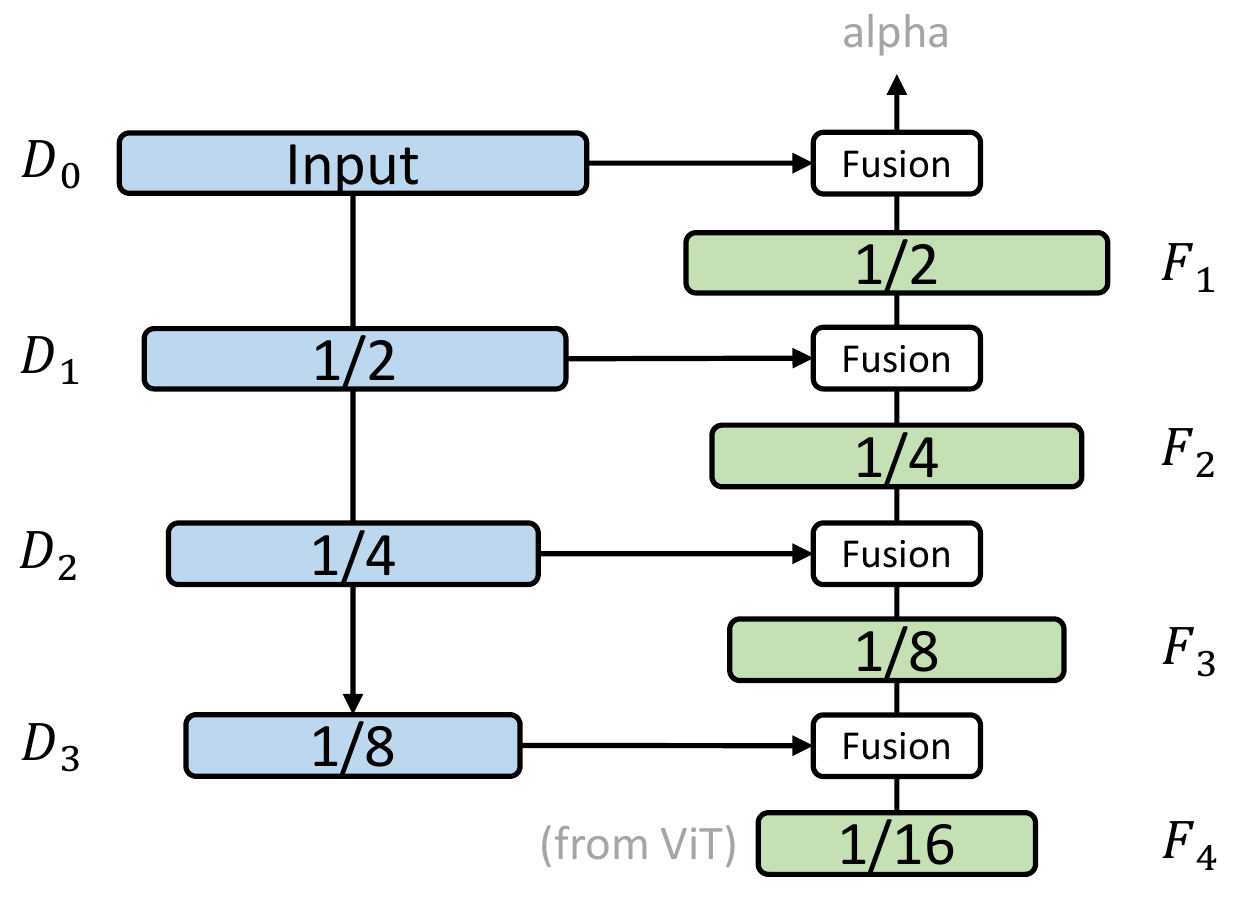}
    \caption{Fuse feature maps output by vision transformers and convstream.}
    \label{fig:fusion}
\end{figure}

As shown in Figure~\ref{fig:fusion}, the output feature map of the ViT is denoted as $F_4$, which is progressively recovered to its original resolution using a simple fusion module. Equation~\eqref{fusison} illustrates the fusion process, which involves upsampling the input feature $F_i$ using bilinear interpolation and concatenating it with the corresponding detail feature map $D_{i-1}$. The resulting feature map is then fused using a simple convolutional layer. By applying the fusion module, $F_4$ is recovered to the original input resolution $\alpha\in R^{H\times W\times 1}$.

\begin{equation}
\label{fusison}
\operatorname{Fusion_i}(F_i, D_{i-1})=\operatorname{Conv}(\operatorname{Upsample}(F_i)\oplus D_{i-1})
\end{equation}

We aim to demonstrate the great potential of ViT in matting tasks, showing that even with a simple decoder, good performance can be achieved when using a strong foundation model. 

\subsection{Training Scheme}
\label{training}

As previously discussed, one of the advantages of \thename{} is its ability to inherit various pretraining weights~\cite{dino, ibot, he2022masked} from ViTs. However, due to the adaptation of the backbone and the trimap input, the vision transformer in \thename{} differs slightly from the original ViT.

To initialize the ViT component in \thename{}, we use pretrained weights from ViT for the original part and randomly initialize the additional parts. Specifically, since our input $cat(X, T)\in R^{H\times W\times 4}$ has 4 channels instead of 3, our patch embedding layer is different from pretrained ones. We employ the FNA++~\cite{fna++} initialization strategy to prevent any performance degradation. This involves mapping the patch-embedding kernels from their original dimensions of $(D, 3, P, P)$ to $(D, 4, P, P)$ by adding zeros to the additional channel, where $D$ represents the embedding dimension and $P$ represents the patch size.

The loss function of \thename{} consists of separate $l_1$ loss, laplacian loss, and gradient penalty loss as follows:

\begin{equation}
\label{total_loss}
\mathcal{L}_{total}=\mathcal{L}_{separate \ l_1}+\mathcal{L}_{lap}+\mathcal{L}_{gp}
\end{equation}

\begin{equation}
\label{separate_l1_loss}
\mathcal{L}_{separate \ l_1}=\frac{1}{|\mathcal{U}|} \sum_{i \in \mathcal{U}}\left|\hat{\alpha}_i-\alpha_i\right| + \frac{1}{|\mathcal{K}|} \sum_{i \in \mathcal{K}}\left|\hat{\alpha}_i-\alpha_i\right|
\end{equation}

To make the network use trimap information for rapid convergence, we do not choose to use l1 loss in the whole image, but calculate l1 loss in the known region and unknown region respectively in trimap. As shown in the equation \eqref{separate_l1_loss}, $\mathcal{U}$ denotes the pixels belonging to the unknown region in the trimap and $\mathcal{K}$ denotes the pixels belonging to the known region in the trimap. Furthermore, to make the network more sensitive to boundaries and get better performance in local smoothness we use laplacian loss as \cite{CAM} and gradient penalty loss as \cite{rmat}.

For data augmentation, we follow the strategy as \cite{GCAMatting}. For an input image, we first randomly affine it including random degree, random scale, random shear, and random flips. Then we randomly crop images to size $512 \times 512$. Random jitter is also applied to change the hue of the image.

\section{Experiments}
Our proposed method, \thename{}, outperforms other existing matting methods and achieves a new \emph{state-of-the-art} performance on the widely-used Composition-1k and Distinctions-646 benchmarks. Furthermore, our \thename{}-S model achieves better results than previous matting methods with even \emph{fewer parameters}.

\begin{figure*}[htp]
  \centering
    \includegraphics[width=1.0\linewidth]{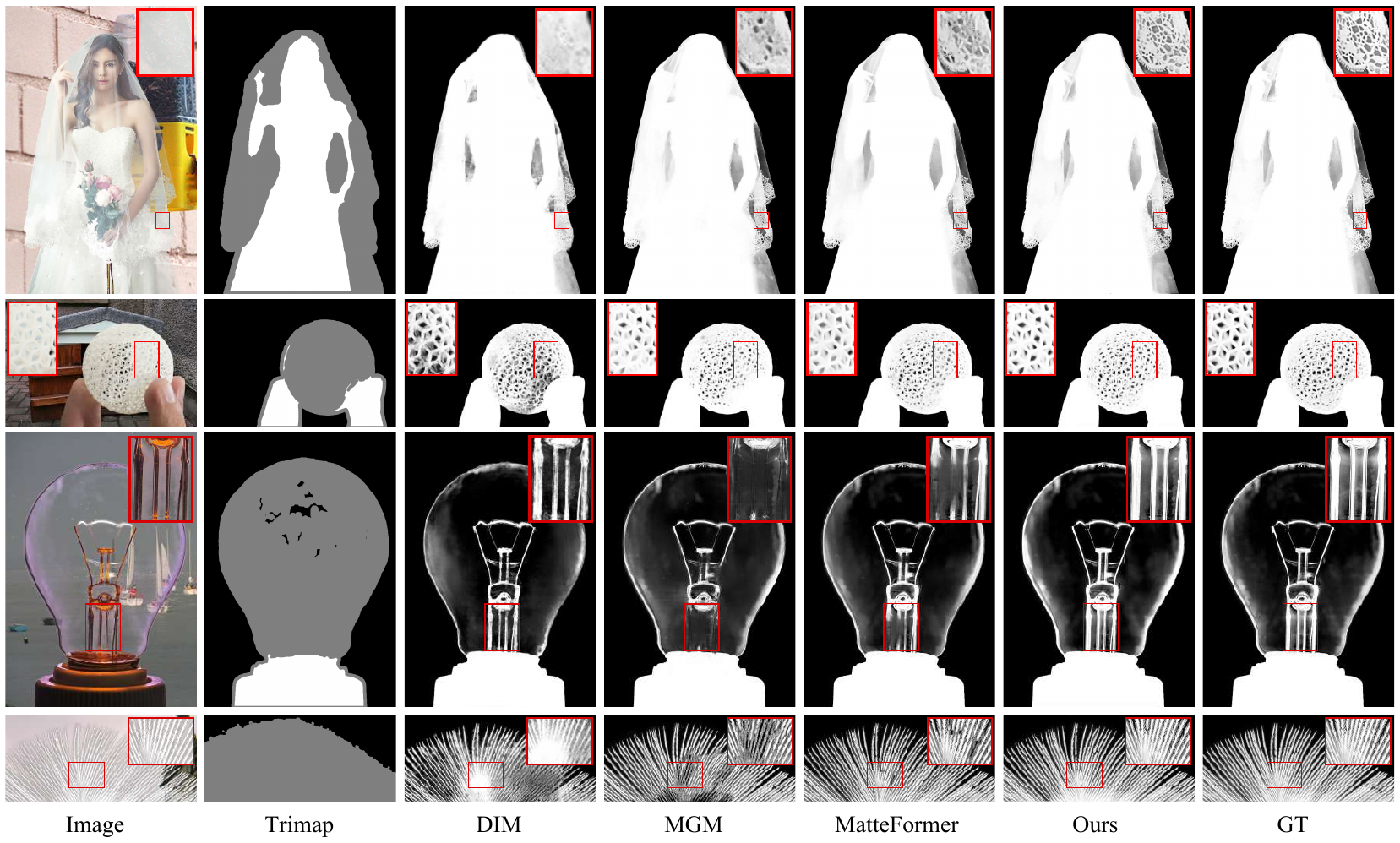}
  \caption{The visual results compared with previous SOTA methods on Composition-1k. Please zoom in for the best view.}
  \label{fig:infer}
\end{figure*}

\subsection{Datasets and Evaluation Metrics}

\textbf{Composition-1k}~\cite{DIM} contains 50 unique foreground images. Each of them is composited with 20 background images from VOC2012~\cite{voc} dataset by $I = \alpha F + (1-\alpha)B$~\cite{rmat} to build the test set with 1000 different synthetic images. The training set, Adobe Image Matting, has 431 unique foreground images. Similar to Composition-1k, each of the foregrounds is composited with 100 background images from COCO~\cite{mscoco} dataset to build a training set containing 43100 different synthetic images.

\textbf{Distinctions-646} is an image matting dataset provided by \cite{HAttMatting}. It uses the same synthetic strategy as Composition-1k. 646 unique foreground images are divided into 596 and 50 images. We build a train set containing 59600 images and a test set containing 1000 images in the same way above.

To evaluate our approach, we use four commonly used metrics: Sum of Absolute Differences (\textbf{SAD}), Mean Square Error (\textbf{MSE}), Gradient loss (\textbf{Grad}), and Connectivity loss (\textbf{Conn}). Lower values of these metrics indicate higher quality of alpha mattes. Note that we scale the value of MSE to 1e-3 for ease of reading.

\subsection{Implementation Details}
For the training data, we first composite our training image in the way mentioned above and generate trimaps by dilation-erosion operation with a random kernel size in $[1, 30]$. We concatenate the RGB image and the trimap and feed it into our model. For the model structure, we build two models \thename{}-S and \thename{}-B in different sizes based on ViT-S and ViT-B backbones~\cite{vit}.  For the model initialization, we initialize \thename-S and \thename-B using DINO~\cite{dino} and MAE~\cite{he2022masked} pretrained weights respectively. We train our model for 100 epochs on two V100 GPUs. For ViTMatte-S and ViTMatte-B, the batchsize is set to 32 and 20 respectively. We use the AdamW optimizer with a learning rate initializing to 5e-4 and weight decay to 0.1.  The learning rate decreased to 0.1 and 0.05 of the original value at epochs 30 and 90. During fine-tuning, a layerwise learning rate is also applied to optimize the pretrained ViT and the decay rate is set to 0.65.

\subsection{Main Results}
\paragraph{Results on Composition-1k} The quantitative results on Composition-1k are shown in Table~\ref{composition-1k}. We measured these metrics only in the unknown regions of trimap. Compared with previous methods, our model outperforms all of them and achieves new \emph{state-of-the-art} (SOTA) performance. Table~\ref{pp-trade-off} compares the performance and number of parameters of our \thename-S with previous SOTA methods. As shown, our method achieves better performance with even a smaller model size. Figure~\ref{fig:infer} shows the qualitative comparison between our approach and previous SOTA methods. For complex cases, \thename{} can effectively capture the details of the image and show excellent performance. 

\paragraph{Results on Distinctions-646} Unlike Composition-1k, Distinctions-646 does not release official trimaps for testing. Previous works typically use randomly generated trimaps for testing, which makes it difficult to do fair comparisons. In our work, we randomly generate trimap according to alpha mattes using erosion operations with random kernel size in $[1, 30]$.  We evaluate the metrics in the whole image region following~\cite{HAttMatting} and the quantitative results are shown in Table~\ref{distinctions-646}. Although the results will be affected by trimaps to some extent, our method still outperforms the previous methods by a large margin, achieving the best results.

\begin{table}[tbp]
\centering
\renewcommand{\arraystretch}{1.1}
\setlength{\tabcolsep}{2mm}{
\begin{tabular}{c|cccc}
    \toprule
    Method                    & SAD$\downarrow$      & MSE$\downarrow$          & Grad$\downarrow$   & Conn$\downarrow$  \\
    \midrule
    Closed-Form~\cite{closed-form}       & 168.1          & 91           & 126.9         & 167.9          \\
    KNN~\cite{knn}               & 175.4          & 103          & 124.1         & 176.4          \\
    DIM~\cite{DIM}                       & 50.4           & 14           & 31.0          & 50.8           \\
    Context-Aware~\cite{CAM}     & 35.8           & 8.2          & 17.3          & 33.2           \\
    A$^2$U~\cite{a2u}                       & 32.2          & 8.2          & 16.4         & 29.3          \\
    MGM~\cite{MGM}                       & 31.5           & 6.8          & 13.5          & 27.3           \\
    SIM~\cite{sim}                       & 28.0           & 5.8          & 10.8          & 24.8           \\
    FBA~\cite{fba}                       & 25.8           & 5.2          & 10.6          & 20.8           \\
    TransMatting~\cite{transmatting}              & 24.96          & 4.58         & 9.72          & 20.16          \\
    MatteFormer~\cite{matteformer}               & 23.80          & 4.03         & 8.68          & 18.90          \\
    RMat~\cite{rmat}                      & 22.87          & 3.9          & 7.74          & 17.84          \\
    \midrule
    ViTMatte-S (Ours) & 21.46 & 3.3 & 7.24 & 16.21 \\
    ViTMatte-B (Ours) & \textbf{20.33} & \textbf{3.0} & \textbf{6.74} & \textbf{14.78} \\
    \bottomrule
\end{tabular}
}
\caption{\textbf{Quantitative results on Composition-1k}.}
\label{composition-1k}
\end{table}

\begin{table}[tbp]
\renewcommand{\arraystretch}{1.1}
\centering
\setlength{\tabcolsep}{2mm}{
\begin{tabular}{c|cccc}
\toprule
Method              & SAD$\downarrow$   & MSE$\downarrow$ & Grad$\downarrow$  & Conn$\downarrow$  \\
\midrule
KNN~\cite{knn}          & 116.68         & 25            & 103.15        & 121.45         \\
DIM~\cite{DIM}                 & 47.56          & 9             & 43.29         & 55.90          \\
HAttMatting$^{*}$ \cite{HAttMatting}         & 48.98          & 9             & 41.57         & 49.93          \\
GCA Matting~\cite{GCAMatting}          & 27.43          & 4.8           & 18.70          & 21.86          \\
MGM~\cite{MGM}          & 33.24          & 4.5          & 20.31         & 25.49          \\
TransMatting~\cite{transmatting}        & 25.65          & 3.4           & 16.08         & 21.45          \\
\midrule
ViTMatte-S(Ours) & 21.22 & 2.1 & 8.78 & 17.55 \\
ViTMatte-B(Ours) & \textbf{17.05}      & \textbf{1.5}     & \textbf{7.03}     & \textbf{12.95}      \\
\bottomrule
\end{tabular}
}
\caption{\textbf{Quantitative results on Distinctions-646}.}
\label{distinctions-646}
\end{table}

\begin{table}[tbp]
\renewcommand{\arraystretch}{1.1}
\centering
\begin{tabular}{c|ccc}
\toprule
Method          & SAD$\downarrow$ &MSE$\downarrow$ & \# params$\downarrow$ \\
\midrule
MatteFormer~\cite{matteformer}               & 23.80          & 4.0          & 44.8M           \\
RMat~\cite{rmat}                      & 22.87          & 3.9          & 27.9M           \\
\midrule
ViTMatte-S(Ours) & 21.46 & 3.3 & \textbf{25.8M}  \\
\bottomrule
\end{tabular}
\caption{\textbf{Comparison of the number of parameters}. \thename{}-S gets better results than previous matting methods with fewer parameters.}
\label{pp-trade-off}
\end{table}

\section{Ablations and Analysis: Comparisons to Previous ViT-based Tasks}

While we are not the first to utilize ViT in downstream vision tasks, our work is influenced by earlier research~\cite{vitdet, vitpose, vitrs} and presents the first direct application of ViT to matting tasks. In this section, we perform a comprehensive structural analysis to investigate the similarities and differences in applying ViT to matting tasks versus other tasks.

By default, we employ ViT-S as the backbone and initialize it with the DINO~\cite{dino} pretrained weights. We train the model for 10 epochs on the Adobe Image Matting~\cite{DIM} dataset and evaluated it on the corresponding benchmark Compositon-1k~\cite{DIM}. To evaluate the performance of each design strategy, we use the Sum of Absolute Differences (SAD) and Mean Squared Error (MSE) metrics.

\subsection{Hybrid Attention for Matting}

Global attention is a powerful tool for the plain vision transformer. Dosovitskiy et al.~\cite{vit} have demonstrated that global attention is highly effective in classification tasks, contributing to the excellent classification performance achieved by the ViT model. However, performing global attention calculations in \emph{each layer} can result in significant computational overheads, which can hinder the scalability of the algorithm for large-scale applications. This raises the question of whether performing global attention calculations at each level is truly necessary for matting tasks.

Empirically, since the trimap already provides sufficient global semantic information, it is expected that the network for matting tasks would be more inclined to focus on detailed information within a small area of the image. This aligns with the mechanism of window attention. Inspired by Li et al.~\cite{vitdet}, we have adopted a novel approach for matting in the \thename{}'s ViT backbone by \emph{alternating between global attention and window attention}, as opposed to computing global attention at each layer, which is done in the original ViT. This new attention mechanism is referred to as \emph{hybrid attention}. Our quantitative and qualitative analysis of this attention mechanism supports our hypothesis.

To quantitatively investigate the impact on performance while reducing computational effort, we replace the global attention in some of the blocks of the ViT backbone with windowed attention. Table~\ref{tab:hybrid} illustrates the superiority of the hybrid attention mechanism in terms of both computational cost and accuracy. Remarkably, replacing the computationally expensive global attention with the less computationally intensive window attention actually results in an increase in network performance. The utilization of four layers of windowed attention reduces computational costs by 50\% compared to full global attention, while maintaining optimal performance, 1.52 improvement on SAD and 0.74 improvement on MSE.

\begin{figure}[tbp]
    \centering
    \includegraphics[width=0.9\linewidth]{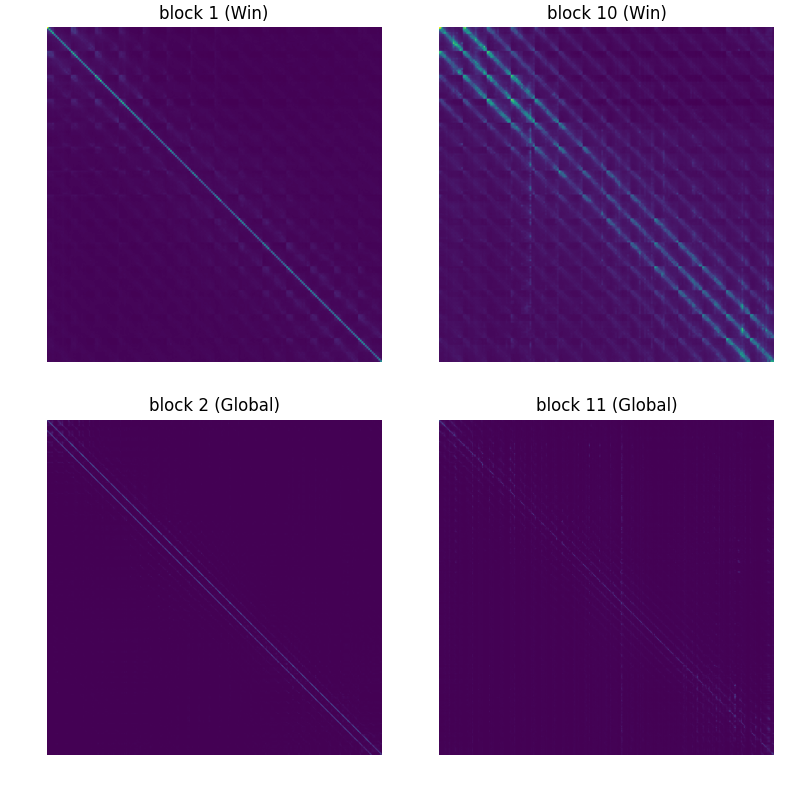}
    \caption{We average and visualize attention maps from window attention blocks and global attention blocks. It reveals that local attention tends to be more strongly activated than global attention.}
    \label{fig:attn_map}
\end{figure}

However, contrary to~\cite{vitdet}, exclusive reliance on global attention does not necessarily yield optimal performance in the matting task. To address this issue, we conduct a qualitative analysis by visualizing the window map of both the window attention and global attention. The resulting attention maps are presented in Figure~\ref{fig:attn_map}. As evident from the figures, the attention map of window attention is more prominently activated in the transformer, whereas the global attention remains in a sub-active state. This observation is in line with our empirical findings, as excessive use of global attention impairs the network's ability to effectively focus on the image details, resulting in degraded performance on image matting.

\begin{table}[tbp]
    \centering
    \renewcommand{\arraystretch}{1.1}
    \begin{tabular}{c|ccc}
        \toprule
        \makecell[c]{num of \\ global attn} & \multicolumn{1}{c}{SAD$\downarrow$} & \multicolumn{1}{c}{MSE$\downarrow$} &\makecell[c]{FLOPs$\downarrow$ \\ (2048$\times$2048)} \\
        \midrule
        12   & 29.83          & 6.38                & 1.00$\times$ (3.28T) \\
        \midrule
        none & 32.30         & 7.03                & 0.26$\times$ \\
        2    & 29.63        & 5.81              & 0.38$\times$ \\
        4    & \textbf{28.31} & \textbf{5.64}       & 0.50$\times$ \\
        8    & 28.40          & 5.82                & 0.63$\times$ \\
        \bottomrule
    \end{tabular}
    \caption{\textbf{Hybrid Attention Mechanism}. Based on it, a plain vision transformer can achieve better matting performance with less computation burden. It helps save about 50\% FLOPs when processing high-resolution images.}
    \label{tab:hybrid}
\end{table}

\begin{table}[tbp]
    \centering
    \renewcommand{\arraystretch}{1.1}
    \begin{tabular}{c|cccc}
    \toprule
        ConvNeck & SAD$\downarrow$ & MSE$\downarrow$ & \# params & \makecell[c]{FLOPs$\downarrow$ \\ (2048$\times$2048)}\\
    \midrule
    none     & 28.31          & 5.64           & 23.9M   & 1.00$\times$ (1.65T) \\
    \midrule
    na\"ive  & 27.99          & 5.34           & 29.2M   & 1.05$\times$ \\
    Residual & \textbf{27.24} & \textbf{5.14}  & 25.8M   & 1.02$\times$ \\
    ConvNeXt & 27.84          & 5.46           & 28.7M   & 1.05$\times$ \\
    \bottomrule
    \end{tabular}
    \caption{\textbf{Performance of different convolution necks}. Convolution neck could improve matting performance of ViTs with negligible FLOPs (2\%-5\%). The residual block performs best in our experiments.}
    \label{convs1}
\end{table}

\subsection{Enhance Transformer Backbones with Convolution Neck}

According to \cite{iformer}, transformers tend to give more attention to low-frequency information. On the other hand, convolution blocks have demonstrated unique advantages in recent studies \cite{xiao2021early, dai2021coatnet}. Incorporating convolutional layers has been shown to improve the performance of image matting when using plain ViT, as well-trained convolution blocks are capable of effectively extracting high-frequency details such as edges and textures~\cite{mimdet}. Successful extraction of such features is a crucial factor in improving the performance of ViT-based models in image matting tasks.

\begin{figure}
    \centering
    \includegraphics[width=0.8\linewidth]{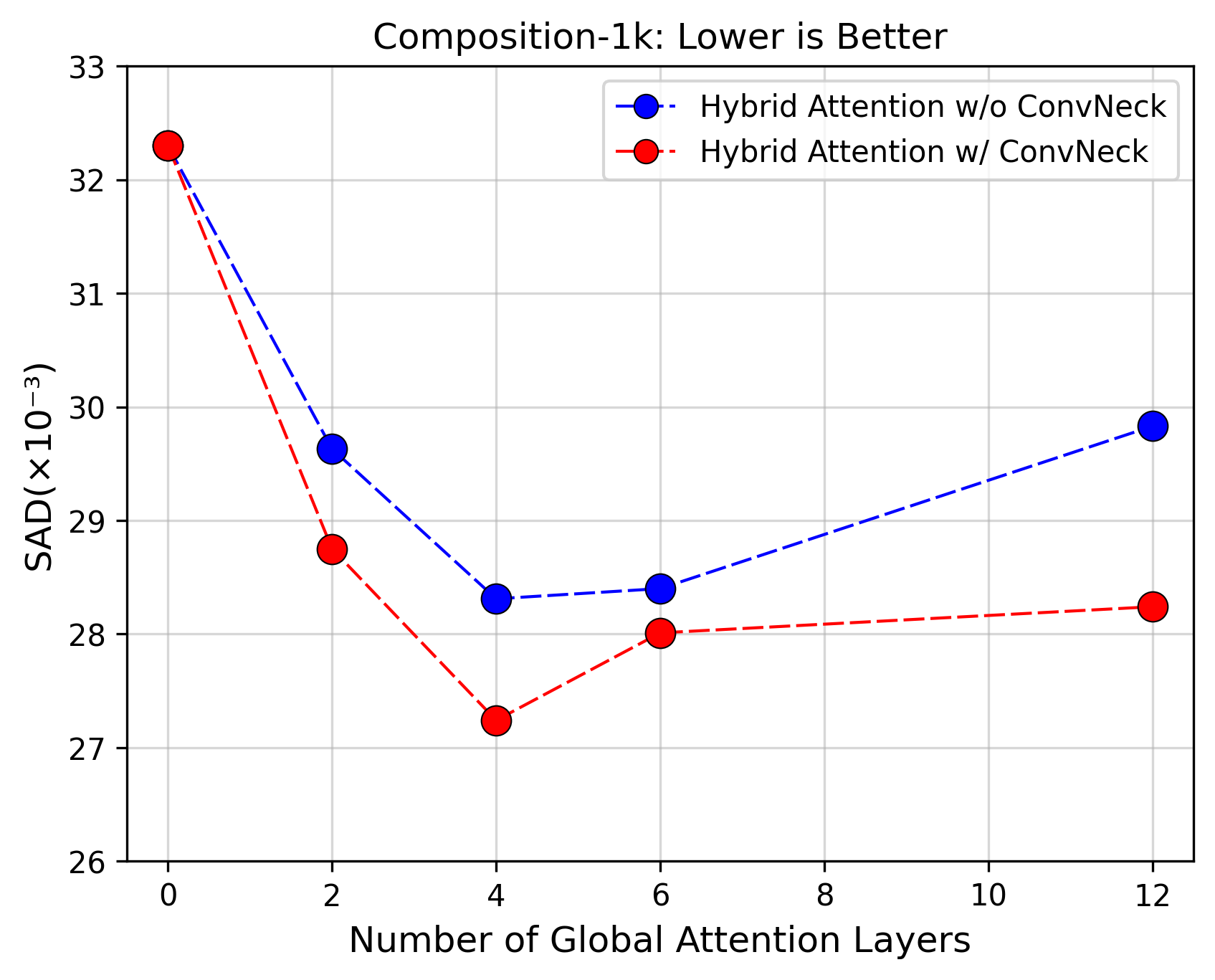}
    \caption{The figure illustrates the improvements we have made to ViT for the matting task, demonstrating that (i) even a lightweight convolutional augmentation of the neck can effectively enhance the model's performance, and (ii) reduce the number of global attention and replacing them with window attentions can further improve the model's performance. This is in line with our observation that the matting task benefits from a design that places greater emphasis on the local details of the image. }
    \label{fig:hybrid_attention}
\end{figure}

Building upon our hybrid attention design, we conduct an experiment where we added a convolutional block after each global attention block. This allows us to investigate the impact of convolution on the overall performance of the network. As shown in Figure~\ref{fig:hybrid_attention}, we observe that the red points were consistently lower than the blue points, indicating that the addition of convolutional blocks contributed significantly to the overall performance of the network. This result further supports our hypothesis and highlights the importance of incorporating convolutional blocks in the architecture of matting networks.

Table~\ref{convs1} presents a comparison of different convolution blocks that can potentially enhance the performance of plain vision transformers for image matting. We fix the number of global attention layers to optimal at 4 and added different convolution blocks after each layer. The three types of convolution blocks used in our experiments are: a na\"ive convolution block with only one 3$\times$3 convolution, a residual bottleneck \cite{res}, and a ConvNeXt block \cite{convnext}. We found that all three convolution blocks improved the matting performance over the baseline with a negligible increase in FLOPs, ranging from 2\% to 5\%. The residual bottleneck performed the best with 1.07 improvement on SAD with only 2\% extra FLOPs, and we incorporated it into our model to enhance its performance.

\subsection{Detail Capture Module}
\label{detail capture module}

\begin{figure}
    \centering
    \includegraphics[width=0.8\linewidth]{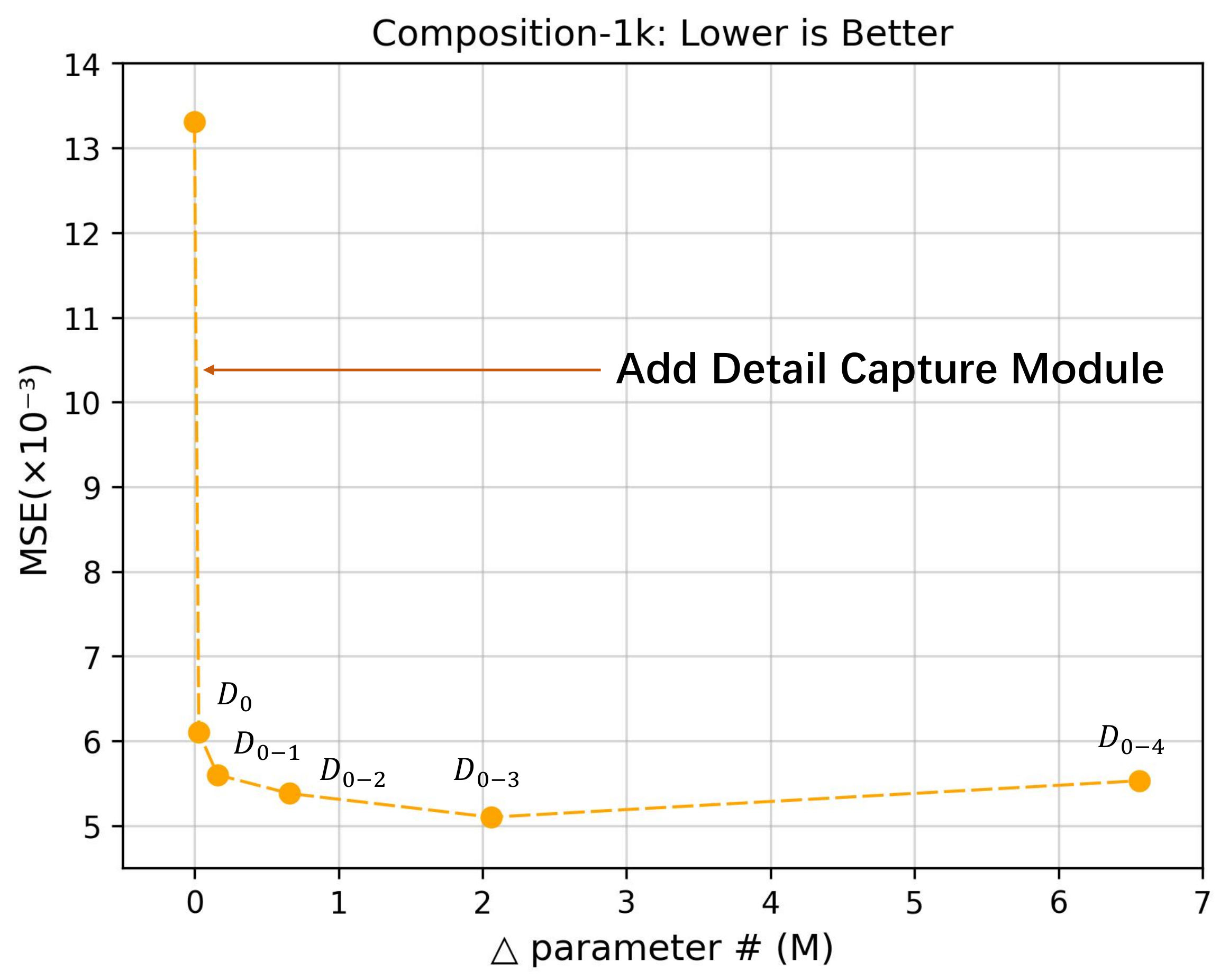}
    \caption{The figure provides evidence for the effectiveness of the Detail Capture Module (DCM), as well as an exploration of the optimal depth for DCM. Experimental results demonstrate that DCM can significantly improve image matting performance with the addition of minimal parameters.}
    \label{fig:DCM}
\end{figure}

The Detail Capture Module (DCM) improves the matting performance significantly. As shown in Figure~\ref{fig:fusion}, we try to change the depth of DCM. For instance, $D_{0-1}$ denotes that we only fuse with detail features $D_0, D_1$. Figure~\ref{fig:DCM} illustrates our results. Adding DCM to the vision transformer with only one feature map could significantly improve 7.21 on MSE performance with less than 0.1M number of parameters. When using $D_0, D_1, D_2$ as discussed in Section~\ref{Detail Capture Module}, \thename{} get the best results.

When using ViT as a backbone for vision tasks,  Simple Feature Pyramid (SFP) introduced by~\cite{vitdet} is a commonly technique to convert the single-scale features of ViT into multi-scale features~\cite{he2022masked, EVA, EVA02}. SFP uses pooling and deconvolution operations to enable ViT to extract features at different levels,  leading to improved performance in handling complex visual scenes. However, SFP may not be well-suited for matting tasks, where high-resolution feature maps are typically necessary to capture fine details~\cite{rmat}. Deconvolution operations used in SFP to obtain high-resolution feature maps can result in significant loss of details and unnecessary computational overhead, limiting their suitability for matting tasks. To address this issue, \thename{} employs a lightweight convstream to extract high-resolution feature maps.

We conduct an experiment to compare these two ways. Specifically, we use independent deconvolutions as ViTDet~\cite{vitdet} to upsample feature map at $\frac{1}{16}$ output by ViT to $\{\frac{1}{2}, \frac{1}{4}, \frac{1}{8} \}$. As shown in Table~\ref{tab:sfp}, our method simultaneously enhances the performance with 1.35 on SAD and 0.40 on MSE while reducing the parameter count by 5.7M of the model. Particularly, when handling high-resolution images, e.g. $(2048, 2048)$, our approach significantly reduces the 71\% computational burden.

\begin{table}[t]
    \centering
    \renewcommand{\arraystretch}{1.2}
    \scalebox{0.7}{
    \begin{tabular}{c|cccc}
    \toprule
        method                      & SAD$\downarrow$                     & MSE$\downarrow$                            & \# params$\downarrow$                           & \makecell[c]{FLOPs$\downarrow$ \\$(2048\times 2048)$} \\
    \midrule
        ViTDet~\cite{vitdet}                      & 28.59                               & 5.57                                       & 31.5M                                        & 5.81T \\
        \thename{}                  & 27.24 (\textcolor{red}{-1.35})    & 5.14 (\textcolor{red}{-0.50})            & 25.8M (\textcolor{red}{-5.7})              & 1.69T (\textcolor{red}{-4.12}) \\
    \bottomrule
    \end{tabular}
    }
    \label{tab:sfp}
    \caption{\textbf{Methods to get multi-scale feature maps}. \thename{} uses DCM instead of SFP used in ViTDet~\cite{vitdet} to get multi-scale features. With DCM, we improve our performance with fewer parameters. It could save about 71\% FLOPs when processing high-resolution images.}
    \label{tab:sfp}
\end{table}

\section{Ablations and Analysis: Comparisons with previous matting systems}

\thename{} is the first study to enhance matting performance using pre-trained ViT models. In the previous section, we demonstrate the superior adaptability of \thename{} over generic ViT-based systems for the matting task. However, a natural question arises: why utilize ViT for matting tasks, and can the application of ViT to matting tasks yield new insights? In this section, we aim to address this question. Specifically, we provide a detailed comparison and analysis of our approach with prior matting methods to demonstrate the novel perspectives that the introduction of ViT brings to matting tasks. 

\subsection{Flexible Pretraining Strategies}

The significance of foundation models in vision tasks is self-evident. These models~\cite{he2022masked, EVA, EVA02, dino, ibot} are typically pretrained using advanced techniques, extensive data, and powerful computing resources. As a result, they possess exceptional capabilities when applied to downstream tasks. However, prior matting methods~\cite{GCAMatting, MGM, matteformer, rmat} often overlooked this aspect. Typically, they simply employed backbones such as ResNet~\cite{res} or SwinTransformer~\cite{swin} pre-trained on ImageNet for the matting task. By contrast, we adopt the highly adaptable ViT structure, which is easily compatible with various powerful pretraining strategies. This means that we can enable the matting task to inherit the advantages of different pretraining strategies without fundamentally altering the model architecture.

\begin{table}[tbp]
    \centering
    \renewcommand{\arraystretch}{1.2}
    \begin{tabular}{c|cc|cc}
    \toprule
                     & \multicolumn{2}{c|}{ViT-S} & \multicolumn{2}{c}{ViT-B} \\
    pretraining      & SAD$\downarrow$  & MSE$\downarrow$ & SAD$\downarrow$  & MSE$\downarrow$ \\
    \midrule
    from scratch     & 59.83          & 19.54         & 50.61  & 16.4 \\
    ImageNet21k      & 29.46          & 6.30          & 26.46  & 5.06 \\
    \textcolor{blue}{MAE}              & /              &/              & \textbf{26.15} &\textbf{4.77} \\
    \textcolor{blue}{DINO}             & \textbf{27.61} &\textbf{5.10}  & 27.89  & 5.22 \\
    \textcolor{blue}{iBOT}             & 28.17          & 5.50          & 28.73  & 5.55 \\
    \bottomrule
    \end{tabular}
    \caption{\textbf{Performace with different pretraining strategies}. \thename{} could inherit different pretraining strategies including supervised and self-supervised pretraining. We observe that \thename{} gets best results with \textcolor{blue}{self-supervised pretraining} DINO~\cite{dino} and MAE~\cite{he2022masked}. }
    \label{pretraining}
\end{table}

As depicted in Table~\ref{pretraining}, we train \thename{} using various pretraining strategies. It is widely known that the convergence of vision transformers heavily relies on large amounts of training data~\cite{vit}. However, since the available matting datasets have limited diversity and size, pretraining a ViT model can potentially alleviate this problem and enhance the overall performance. As shown in Table~\ref{pretraining}, training \thename{} from scratch results in significant performance degradation. Nevertheless, \thename{} can easily inherit different pretraining strategies, including both supervised and self-supervised pretraining. Our experiments reveal that ViT models with self-supervised pretraining, such as MAE~\cite{he2022masked} and DINO~\cite{dino}, yield the best performance for \thename{}. This observation highlights the superiority of our method over other matting techniques. In the future, better pretrained weights may consistently improve matting performance via ViTMatte.

\subsection{Lightest-weight Decoder}

\begin{figure}
    \centering
    \includegraphics[width=0.9\linewidth]{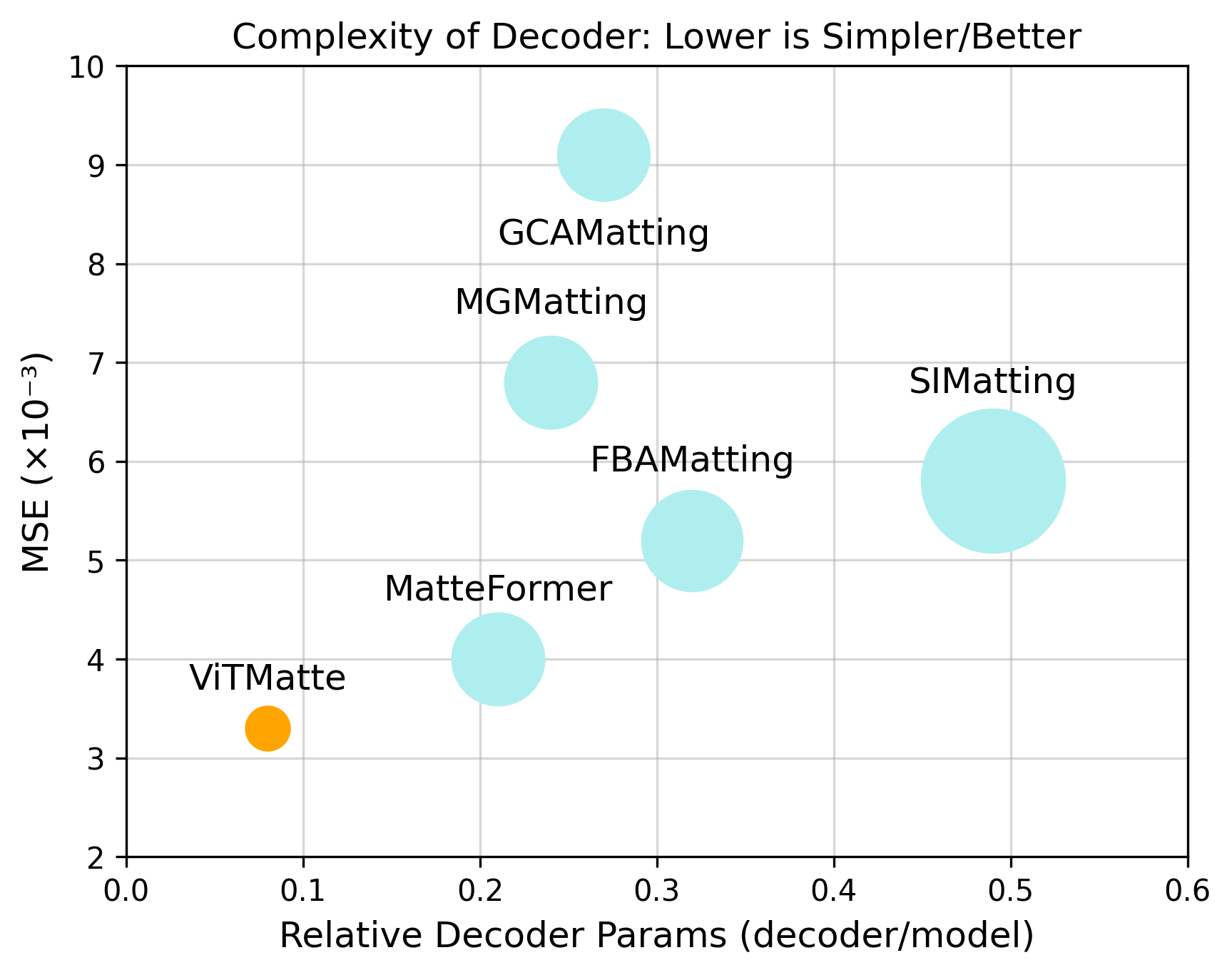}
    \caption{We compare the size and complexity of the decoder in our approach with that of existing matting methods. Remarkably, \thename{} has the smallest and simplest decoder yet achieves the best matting performance. This suggests that a sophisticated decoder design may \emph{not} be necessary for matting methods when using a ``foundation" model.}
    \label{fig:matting_decoders}
\end{figure}

\begin{figure*}[htp]
  \centering
    \includegraphics[width=1.0\linewidth]{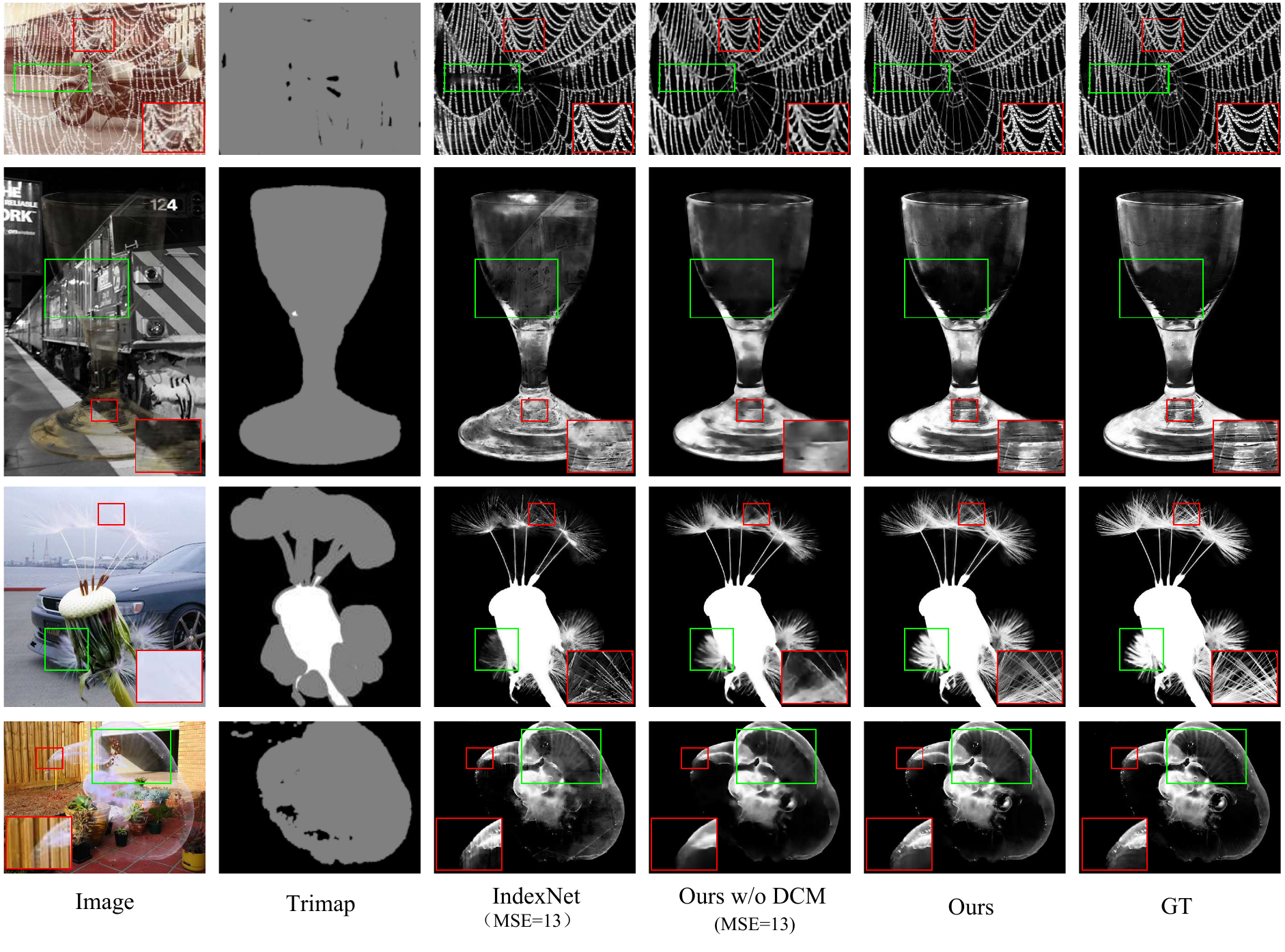}
  \caption{Visual Results of IndexNet, \thename{} \emph{without DCM} and \thename{}. Please zoom in for the best view.}
  \label{fig:decoder_infer}
\end{figure*}

\thename{} adopts the lightest matting decoder ever. While other methods are pursuing more complicated decoders tailored for matting, \thename{} pays more attention to the foundation part of the design. In \thename{}, we treat all of the Detail Capture Module as our decoder. As discussed in Section~\ref{detail capture module}, we use DCM to simplify the most commonly used Simple Feature Pyramid \emph{ViT-based} tasks. However, is it still superior to other \emph{matting} methods? 

To provide more insight, we conduct a quantitative comparison between our method and other recent approaches \cite{matteformer, fba, sim, MGM, GCAMatting}. In addition to accuracy (measured in terms of MSE), we also assess the number of encoder parameters, the number of parameters in the overall model, as well as their ratio denoted as \emph{Relative Decoder Params} in Figure~\ref{fig:matting_decoders}. As shown, our method achieves the highest accuracy while employing the smallest number of decoder parameters. Moreover, it is essential to note that the ratio of the decoder in our model is also the smallest (8\%), which is significantly smaller than previous approaches (ranging from 21\% to 49\%). These results effectively address the question raised in the main paper and suggest that the plain ViT backbone plays the most critical role in image matting. Furthermore, this finding reinforces the emerging paradigm of separating task-agnostic pretraining from task-specific lightweight adaptation. In other words, we can conclude that a decoder with high inductive bias (as used in prior methods) may not be necessary for image matting.

We visualize the results and compare the impact of the Detail Capture Module (DCM) on the visual performance of the model. Our \thename{} \emph{without DCM}  has similar quantitative results with IndexNet\cite{indexnet}. However, they have obvious differences in visualization.

Figure~\ref{fig:decoder_infer} illustrates the impact of the Detail Capture Module on the performance of \thename{}. Although the model without the decoder can produce visually plausible alpha mattes, a closer examination of the results in the zoomed-in regions of the \textcolor{red}{red boxes} reveals that they tend to be over-smoothed and lack expressive visual details compared to the other two groups. On the other hand, both variants of \thename{} equipped with the decoder show more visual details. Furthermore, the \textcolor{green}{green boxes} indicate that \thename{} without the decoder does not suffer from obvious semantic errors, such as background mapping or region loss, which are present in the IndexNet~\cite{indexnet} approach. These visual comparisons demonstrate that the Detail Capture Module can effectively capture and incorporate detailed information for matting, while the ViT backbone provides the major computational power to solve the matting problem.

    

\subsection{Flexible Inference Strategies}

\begin{table}[htp]
\centering
\begin{tabular}{c|ccc}
\toprule
\begin{tabular}[c]{@{}c@{}}infer\\ strategy\end{tabular} & SAD$\downarrow$   & \begin{tabular}[c]{@{}c@{}}MSE$\downarrow$\\ ($\times 10^{-3}$)\end{tabular}  & \begin{tabular}[c]{@{}c@{}}infer\\ mem.\end{tabular} \\
\midrule
normal                                                   & 21.46 & 3.28 & 20.3G                                               \\
grid sample                                              & 21.54 & 3.34 & \textbf{6.8G} (\textcolor{red}{-66\%}) \\
\bottomrule
\end{tabular}
\caption{\textbf{Grid sample inference}. Grid sample could effectively reduce memory burden with negligible performance drop while inferring high-resolution images. \emph{Infer mem.} is tested on images with size (2048, 2048).}
\label{tab_grid_sample}
\end{table}

\begin{figure}
    \centering
    \includegraphics[width=0.8\linewidth]{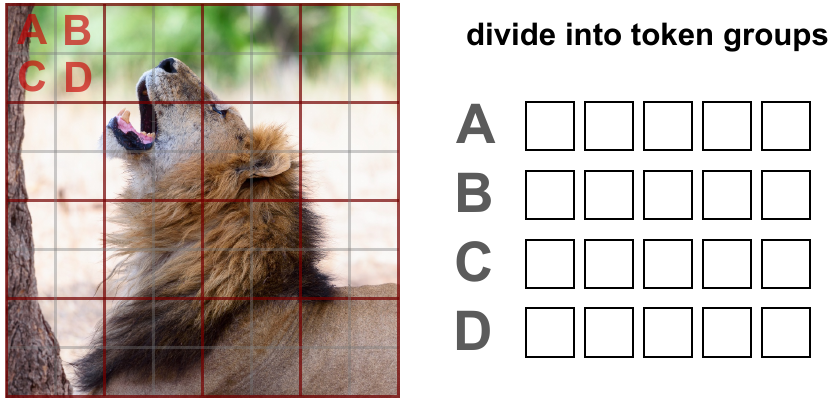}
    \caption{Grid sample inference.}
    \label{fig:grid_sample}
\end{figure}

Thanks to our backbone adaptation strategy, training cost is effectively reduced and our \thename{} can be applied to most scenarios. However, when facing high-resolution images, ViT's global attention still introduces a high computation burden. To solve the problem, we employ a simple inference strategy, as shown in Figure~\ref{fig:grid_sample}. Specifically, we grid-sample the tokens before global attention. Each grid contains 4 image tokens denoted as $A, B, C, D$. We divide the tokens of all girds into four groups and calculate the self-attention for each group of tokens.

As shown in Table~\ref{tab_grid_sample}, we use different strategies while inferring on Composition-1k~\cite{DIM}. Notably that the grid sample is only while inferring, the training strategy is the same as discussed in the main text. Surprisingly, it saves a lot of memory of GPU while inferring at a slight performance penalty. Our \thename{} demonstrates strong flexibility in inference.

\section{Conclusion}

In this paper, we present a concise and efficient matting system based on plain vision transformers, named \thename{}. We use a hybrid mechanism and a convolution neck to adapt ViT for image matting. Besides, we also design the lightest detail capture module as a decoder among previous matting methods to complement the detailed information required by matting. For the first time, we demonstrate the great potential of pretrained ViT on image matting tasks. We compare it to the previous ViT adaptation strategy and demonstrate the superiority of \thename{} in the matting task. We also compare it to previous matting systems and find our method has various unique advantages, such as concise structure and various pretraining strategies. Benefiting from the rapid development of ViT-based vision foundation models,  we hope \thename{} would be a standard tool in matting-related industrial applications.

{\small
\bibliographystyle{ieee_fullname}
\bibliography{ref}

\begin{thebibliography}{10}\itemsep=-1pt

\bibitem{matting_survey}
Jagruti Boda and Dhatri Pandya.
\newblock A survey on image matting techniques.
\newblock In {\em 2018 International Conference on Communication and Signal
  Processing (ICCSP)}, pages 0765--0770, 2018.

\bibitem{bommasani2021opportunities}
Rishi Bommasani, Drew~A Hudson, Ehsan Adeli, Russ Altman, Simran Arora, Sydney
  von Arx, Michael~S Bernstein, Jeannette Bohg, Antoine Bosselut, Emma
  Brunskill, et~al.
\newblock On the opportunities and risks of foundation models.
\newblock {\em arXiv preprint arXiv:2108.07258}, 2021.

\bibitem{GPT3}
Tom Brown, Benjamin Mann, Nick Ryder, Melanie Subbiah, Jared~D Kaplan, Prafulla
  Dhariwal, Arvind Neelakantan, Pranav Shyam, Girish Sastry, Amanda Askell,
  Sandhini Agarwal, Ariel Herbert-Voss, Gretchen Krueger, Tom Henighan, Rewon
  Child, Aditya Ramesh, Daniel Ziegler, Jeffrey Wu, Clemens Winter, Chris
  Hesse, Mark Chen, Eric Sigler, Mateusz Litwin, Scott Gray, Benjamin Chess,
  Jack Clark, Christopher Berner, Sam McCandlish, Alec Radford, Ilya Sutskever,
  and Dario Amodei.
\newblock Language models are few-shot learners.
\newblock In H. Larochelle, M. Ranzato, R. Hadsell, M.F. Balcan, and H. Lin,
  editors, {\em Advances in Neural Information Processing Systems}, volume~33,
  pages 1877--1901. Curran Associates, Inc., 2020.

\bibitem{brown2020language}
Tom Brown, Benjamin Mann, Nick Ryder, Melanie Subbiah, Jared~D Kaplan, Prafulla
  Dhariwal, Arvind Neelakantan, Pranav Shyam, Girish Sastry, Amanda Askell,
  et~al.
\newblock Language models are few-shot learners.
\newblock {\em Advances in neural information processing systems},
  33:1877--1901, 2020.

\bibitem{transmatting}
Huanqia Cai, Fanglei Xue, Lele Xu, and Lili Guo.
\newblock Transmatting: Enhancing transparent objects matting with
  transformers.
\newblock In {\em European Conference on Computer Vision}, pages 253--269.
  Springer, 2022.

\bibitem{dino}
Mathilde Caron, Hugo Touvron, Ishan Misra, Herv{\'e} J{\'e}gou, Julien Mairal,
  Piotr Bojanowski, and Armand Joulin.
\newblock Emerging properties in self-supervised vision transformers.
\newblock In {\em Proceedings of the IEEE/CVF International Conference on
  Computer Vision}, pages 9650--9660, 2021.

\bibitem{knnmatting2013}
Qifeng Chen, Dingzeyu Li, and Chi-Keung Tang.
\newblock Knn matting.
\newblock {\em IEEE Transactions on Pattern Analysis and Machine Intelligence},
  35(9):2175--2188, 2013.

\bibitem{knn}
Qifeng Chen, Dingzeyu Li, and Chi-Keung Tang.
\newblock Knn matting.
\newblock {\em IEEE transactions on pattern analysis and machine intelligence},
  35(9):2175--2188, 2013.

\bibitem{a2u}
Yutong Dai, Hao Lu, and Chunhua Shen.
\newblock Learning affinity-aware upsampling for deep image matting.
\newblock In {\em Proceedings of the IEEE/CVF Conference on Computer Vision and
  Pattern Recognition}, pages 6841--6850, 2021.

\bibitem{rmat}
Yutong Dai, Brian Price, He Zhang, and Chunhua Shen.
\newblock Boosting robustness of image matting with context assembling and
  strong data augmentation.
\newblock In {\em Proceedings of the IEEE/CVF Conference on Computer Vision and
  Pattern Recognition}, pages 11707--11716, 2022.

\bibitem{dai2021coatnet}
Zihang Dai, Hanxiao Liu, Quoc~V Le, and Mingxing Tan.
\newblock Coatnet: Marrying convolution and attention for all data sizes.
\newblock {\em Advances in Neural Information Processing Systems},
  34:3965--3977, 2021.

\bibitem{deng2009imagenet}
Jia Deng, Wei Dong, Richard Socher, Li-Jia Li, Kai Li, and Li Fei-Fei.
\newblock Imagenet: A large-scale hierarchical image database.
\newblock In {\em 2009 IEEE conference on computer vision and pattern
  recognition}, pages 248--255. Ieee, 2009.

\bibitem{devlin2018bert}
Jacob Devlin, Ming-Wei Chang, Kenton Lee, and Kristina Toutanova.
\newblock Bert: Pre-training of deep bidirectional transformers for language
  understanding.
\newblock {\em arXiv preprint arXiv:1810.04805}, 2018.

\bibitem{vit}
Alexey Dosovitskiy, Lucas Beyer, Alexander Kolesnikov, Dirk Weissenborn,
  Xiaohua Zhai, Thomas Unterthiner, Mostafa Dehghani, Matthias Minderer, Georg
  Heigold, Sylvain Gelly, et~al.
\newblock An image is worth 16x16 words: Transformers for image recognition at
  scale.
\newblock {\em arXiv preprint arXiv:2010.11929}, 2020.

\bibitem{voc}
Mark Everingham, Luc Van~Gool, Christopher~KI Williams, John Winn, and Andrew
  Zisserman.
\newblock The pascal visual object classes (voc) challenge.
\newblock {\em International journal of computer vision}, 88(2):303--338, 2010.

\bibitem{fan2021multiscale}
Haoqi Fan, Bo Xiong, Karttikeya Mangalam, Yanghao Li, Zhicheng Yan, Jitendra
  Malik, and Christoph Feichtenhofer.
\newblock Multiscale vision transformers.
\newblock In {\em Proceedings of the IEEE/CVF International Conference on
  Computer Vision}, pages 6824--6835, 2021.

\bibitem{fna++}
Jiemin Fang, Yuzhu Sun, Qian Zhang, Kangjian Peng, Yuan Li, Wenyu Liu, and
  Xinggang Wang.
\newblock Fna++: Fast network adaptation via parameter remapping and
  architecture search.
\newblock {\em IEEE Transactions on Pattern Analysis and Machine Intelligence},
  43(9):2990--3004, 2020.

\bibitem{EVA02}
Yuxin Fang, Quan Sun, Xinggang Wang, Tiejun Huang, Xinlong Wang, and Yue Cao.
\newblock Eva-02: A visual representation for neon genesis.
\newblock {\em arXiv preprint arXiv:2303.11331}, 2023.

\bibitem{EVA}
Yuxin Fang, Wen Wang, Binhui Xie, Quan Sun, Ledell Wu, Xinggang Wang, Tiejun
  Huang, Xinlong Wang, and Yue Cao.
\newblock Eva: Exploring the limits of masked visual representation learning at
  scale.
\newblock {\em arXiv preprint arXiv:2211.07636}, 2022.

\bibitem{mimdet}
Yuxin Fang, Shusheng Yang, Shijie Wang, Yixiao Ge, Ying Shan, and Xinggang
  Wang.
\newblock Unleashing vanilla vision transformer with masked image modeling for
  object detection.
\newblock {\em arXiv preprint arXiv:2204.02964}, 2022.

\bibitem{fba}
Marco Forte and Fran{\c{c}}ois Piti{\'e}.
\newblock $ f $, $ b $, alpha matting.
\newblock {\em arXiv preprint arXiv:2003.07711}, 2020.

\bibitem{he2022masked}
Kaiming He, Xinlei Chen, Saining Xie, Yanghao Li, Piotr Doll{\'a}r, and Ross
  Girshick.
\newblock Masked autoencoders are scalable vision learners.
\newblock In {\em Proceedings of the IEEE/CVF Conference on Computer Vision and
  Pattern Recognition}, pages 16000--16009, 2022.

\bibitem{hematting2011}
Kaiming He, Christoph Rhemann, Carsten Rother, Xiaoou Tang, and Jian Sun.
\newblock A global sampling method for alpha matting.
\newblock In {\em CVPR 2011}, pages 2049--2056, 2011.

\bibitem{res}
Kaiming He, Xiangyu Zhang, Shaoqing Ren, and Jian Sun.
\newblock Deep residual learning for image recognition.
\newblock In {\em Proceedings of the IEEE conference on computer vision and
  pattern recognition}, pages 770--778, 2016.

\bibitem{heo2021rethinking}
Byeongho Heo, Sangdoo Yun, Dongyoon Han, Sanghyuk Chun, Junsuk Choe, and
  Seong~Joon Oh.
\newblock Rethinking spatial dimensions of vision transformers.
\newblock In {\em Proceedings of the IEEE/CVF International Conference on
  Computer Vision}, pages 11936--11945, 2021.

\bibitem{CAM}
Qiqi Hou and Feng Liu.
\newblock Context-aware image matting for simultaneous foreground and alpha
  estimation.
\newblock In {\em Proceedings of the IEEE/CVF International Conference on
  Computer Vision}, pages 4130--4139, 2019.

\bibitem{nonlocalmatting2011}
Philip Lee and Ying Wu.
\newblock Nonlocal matting.
\newblock In {\em CVPR 2011}, pages 2193--2200, 2011.

\bibitem{closed-form}
Anat Levin, Dani Lischinski, and Yair Weiss.
\newblock A closed-form solution to natural image matting.
\newblock {\em IEEE transactions on pattern analysis and machine intelligence},
  30(2):228--242, 2007.

\bibitem{aim500}
Jizhizi Li, Jing Zhang, and Dacheng Tao.
\newblock Deep automatic natural image matting.
\newblock In Zhi-Hua Zhou, editor, {\em Proceedings of the Thirtieth
  International Joint Conference on Artificial Intelligence, {IJCAI-21}}, pages
  800--806. International Joint Conferences on Artificial Intelligence
  Organization, 8 2021.
\newblock Main Track.

\bibitem{GCAMatting}
Yaoyi Li and Hongtao Lu.
\newblock Natural image matting via guided contextual attention.
\newblock In {\em Proceedings of the AAAI Conference on Artificial
  Intelligence}, volume~34, pages 11450--11457, 2020.

\bibitem{vitdet}
Yanghao Li, Hanzi Mao, Ross Girshick, and Kaiming He.
\newblock Exploring plain vision transformer backbones for object detection.
\newblock {\em arXiv preprint arXiv:2203.16527}, 2022.

\bibitem{rvm}
Shanchuan Lin, Linjie Yang, Imran Saleemi, and Soumyadip Sengupta.
\newblock Robust high-resolution video matting with temporal guidance, 2021.

\bibitem{Lin_2022_WACV}
Shanchuan Lin, Linjie Yang, Imran Saleemi, and Soumyadip Sengupta.
\newblock Robust high-resolution video matting with temporal guidance.
\newblock In {\em Proceedings of the IEEE/CVF Winter Conference on Applications
  of Computer Vision (WACV)}, pages 238--247, January 2022.

\bibitem{fpn}
Tsung-Yi Lin, Piotr Doll{\'a}r, Ross Girshick, Kaiming He, Bharath Hariharan,
  and Serge Belongie.
\newblock Feature pyramid networks for object detection.
\newblock In {\em Proceedings of the IEEE conference on computer vision and
  pattern recognition}, pages 2117--2125, 2017.

\bibitem{mscoco}
Tsung-Yi Lin, Michael Maire, Serge Belongie, James Hays, Pietro Perona, Deva
  Ramanan, Piotr Doll{\'a}r, and C~Lawrence Zitnick.
\newblock Microsoft coco: Common objects in context.
\newblock In {\em European conference on computer vision}, pages 740--755.
  Springer, 2014.

\bibitem{Tripartitle2021}
Yuhao Liu, Jiake Xie, Xiao Shi, Yu Qiao, Yujie Huang, Yong Tang, and Xin Yang.
\newblock Tripartite information mining and integration for image matting.
\newblock In {\em 2021 IEEE/CVF International Conference on Computer Vision
  (ICCV)}, pages 7535--7544, 2021.

\bibitem{swin}
Ze Liu, Yutong Lin, Yue Cao, Han Hu, Yixuan Wei, Zheng Zhang, Stephen Lin, and
  Baining Guo.
\newblock Swin transformer: Hierarchical vision transformer using shifted
  windows.
\newblock In {\em Proceedings of the IEEE/CVF International Conference on
  Computer Vision}, pages 10012--10022, 2021.

\bibitem{convnext}
Zhuang Liu, Hanzi Mao, Chao-Yuan Wu, Christoph Feichtenhofer, Trevor Darrell,
  and Saining Xie.
\newblock A convnet for the 2020s.
\newblock In {\em Proceedings of the IEEE/CVF Conference on Computer Vision and
  Pattern Recognition}, pages 11976--11986, 2022.

\bibitem{indexnet}
Hao Lu, Yutong Dai, Chunhua Shen, and Songcen Xu.
\newblock Indices matter: Learning to index for deep image matting.
\newblock In {\em Proceedings of the IEEE/CVF International Conference on
  Computer Vision}, pages 3266--3275, 2019.

\bibitem{matteformer}
GyuTae Park, SungJoon Son, JaeYoung Yoo, SeHo Kim, and Nojun Kwak.
\newblock Matteformer: Transformer-based image matting via prior-tokens.
\newblock In {\em Proceedings of the IEEE/CVF Conference on Computer Vision and
  Pattern Recognition}, pages 11696--11706, 2022.

\bibitem{HAttMatting}
Yu Qiao, Yuhao Liu, Xin Yang, Dongsheng Zhou, Mingliang Xu, Qiang Zhang, and
  Xiaopeng Wei.
\newblock Attention-guided hierarchical structure aggregation for image
  matting.
\newblock In {\em Proceedings of the IEEE/CVF Conference on Computer Vision and
  Pattern Recognition}, pages 13676--13685, 2020.

\bibitem{radford2018improving}
Alec Radford, Karthik Narasimhan, Tim Salimans, Ilya Sutskever, et~al.
\newblock Improving language understanding by generative pre-training.
\newblock 2018.

\bibitem{radford2019language}
Alec Radford, Jeffrey Wu, Rewon Child, David Luan, Dario Amodei, Ilya
  Sutskever, et~al.
\newblock Language models are unsupervised multitask learners.
\newblock {\em OpenAI blog}, 1(8):9, 2019.

\bibitem{matting2013}
Ehsan Shahrian, Deepu Rajan, Brian Price, and Scott Cohen.
\newblock Improving image matting using comprehensive sampling sets.
\newblock In {\em 2013 IEEE Conference on Computer Vision and Pattern
  Recognition}, pages 636--643, 2013.

\bibitem{iformer}
Chenyang Si, Weihao Yu, Pan Zhou, Yichen Zhou, Xinchao Wang, and Shuicheng Yan.
\newblock Inception transformer.
\newblock {\em arXiv preprint arXiv:2205.12956}, 2022.

\bibitem{poissonmatting04}
Jian Sun, Jiaya Jia, Chi-Keung Tang, and Heung-Yeung Shum.
\newblock Poisson matting.
\newblock 23(3), 2004.

\bibitem{sim}
Yanan Sun, Chi-Keung Tang, and Yu-Wing Tai.
\newblock Semantic image matting.
\newblock In {\em Proceedings of the IEEE/CVF Conference on Computer Vision and
  Pattern Recognition}, pages 11120--11129, 2021.

\bibitem{Tang_2019_CVPR}
Jingwei Tang, Yagiz Aksoy, Cengiz Oztireli, Markus Gross, and Tunc~Ozan Aydin.
\newblock Learning-based sampling for natural image matting.
\newblock In {\em Proceedings of the IEEE/CVF Conference on Computer Vision and
  Pattern Recognition (CVPR)}, June 2019.

\bibitem{attention}
Ashish Vaswani, Noam Shazeer, Niki Parmar, Jakob Uszkoreit, Llion Jones,
  Aidan~N Gomez, {\L}ukasz Kaiser, and Illia Polosukhin.
\newblock Attention is all you need.
\newblock {\em Advances in neural information processing systems}, 30, 2017.

\bibitem{NIPS2017_3f5ee243}
Ashish Vaswani, Noam Shazeer, Niki Parmar, Jakob Uszkoreit, Llion Jones,
  Aidan~N Gomez, \L~ukasz Kaiser, and Illia Polosukhin.
\newblock Attention is all you need.
\newblock In I. Guyon, U.~Von Luxburg, S. Bengio, H. Wallach, R. Fergus, S.
  Vishwanathan, and R. Garnett, editors, {\em Advances in Neural Information
  Processing Systems}, volume~30. Curran Associates, Inc., 2017.

\bibitem{vitrs}
Di Wang, Qiming Zhang, Yufei Xu, Jing Zhang, Bo Du, Dacheng Tao, and Liangpei
  Zhang.
\newblock Advancing plain vision transformer towards remote sensing foundation
  model.
\newblock {\em IEEE Transactions on Geoscience and Remote Sensing}, 2022.

\bibitem{Wang_2021_ICCV}
Tiantian Wang, Sifei Liu, Yapeng Tian, Kai Li, and Ming-Hsuan Yang.
\newblock Video matting via consistency-regularized graph neural networks.
\newblock In {\em Proceedings of the IEEE/CVF International Conference on
  Computer Vision (ICCV)}, pages 4902--4911, October 2021.

\bibitem{PVT}
Wenhai Wang, Enze Xie, Xiang Li, Deng-Ping Fan, Kaitao Song, Ding Liang, Tong
  Lu, Ping Luo, and Ling Shao.
\newblock Pyramid vision transformer: A versatile backbone for dense prediction
  without convolutions.
\newblock In {\em Proceedings of the IEEE/CVF International Conference on
  Computer Vision}, pages 568--578, 2021.

\bibitem{xiao2021early}
Tete Xiao, Mannat Singh, Eric Mintun, Trevor Darrell, Piotr Doll{\'a}r, and
  Ross Girshick.
\newblock Early convolutions help transformers see better.
\newblock {\em Advances in Neural Information Processing Systems},
  34:30392--30400, 2021.

\bibitem{segformer}
Enze Xie, Wenhai Wang, Zhiding Yu, Anima Anandkumar, Jose~M Alvarez, and Ping
  Luo.
\newblock Segformer: Simple and efficient design for semantic segmentation with
  transformers.
\newblock {\em Advances in Neural Information Processing Systems},
  34:12077--12090, 2021.

\bibitem{DIM}
Ning Xu, Brian Price, Scott Cohen, and Thomas Huang.
\newblock Deep image matting.
\newblock In {\em Proceedings of the IEEE conference on computer vision and
  pattern recognition}, pages 2970--2979, 2017.

\bibitem{vitpose}
Yufei Xu, Jing Zhang, Qiming Zhang, and Dacheng Tao.
\newblock Vitpose: Simple vision transformer baselines for human pose
  estimation.
\newblock {\em arXiv preprint arXiv:2204.12484}, 2022.

\bibitem{HDMatt}
Haichao Yu, Ning Xu, Zilong Huang, Yuqian Zhou, and Humphrey Shi.
\newblock High-resolution deep image matting.
\newblock In {\em Proceedings of the AAAI Conference on Artificial
  Intelligence}, volume~35, pages 3217--3224, 2021.

\bibitem{MGM}
Qihang Yu, Jianming Zhang, He Zhang, Yilin Wang, Zhe Lin, Ning Xu, Yutong Bai,
  and Alan Yuille.
\newblock Mask guided matting via progressive refinement network.
\newblock In {\em Proceedings of the IEEE/CVF Conference on Computer Vision and
  Pattern Recognition}, pages 1154--1163, 2021.

\bibitem{Florence}
Lu Yuan, Dongdong Chen, Yi{-}Ling Chen, Noel Codella, Xiyang Dai, Jianfeng Gao,
  Houdong Hu, Xuedong Huang, Boxin Li, Chunyuan Li, Ce Liu, Mengchen Liu,
  Zicheng Liu, Yumao Lu, Yu Shi, Lijuan Wang, Jianfeng Wang, Bin Xiao, Zhen
  Xiao, Jianwei Yang, Michael Zeng, Luowei Zhou, and Pengchuan Zhang.
\newblock Florence: {A} new foundation model for computer vision.
\newblock {\em CoRR}, abs/2111.11432, 2021.

\bibitem{T2T}
Li Yuan, Yunpeng Chen, Tao Wang, Weihao Yu, Yujun Shi, Zi-Hang Jiang,
  Francis~EH Tay, Jiashi Feng, and Shuicheng Yan.
\newblock Tokens-to-token vit: Training vision transformers from scratch on
  imagenet.
\newblock In {\em Proceedings of the IEEE/CVF International Conference on
  Computer Vision}, pages 558--567, 2021.

\bibitem{Zhang_2019_CVPR}
Yunke Zhang, Lixue Gong, Lubin Fan, Peiran Ren, Qixing Huang, Hujun Bao, and
  Weiwei Xu.
\newblock A late fusion cnn for digital matting.
\newblock In {\em Proceedings of the IEEE/CVF Conference on Computer Vision and
  Pattern Recognition (CVPR)}, June 2019.

\bibitem{ibot}
Jinghao Zhou, Chen Wei, Huiyu Wang, Wei Shen, Cihang Xie, Alan Yuille, and Tao
  Kong.
\newblock ibot: Image bert pre-training with online tokenizer.
\newblock {\em arXiv preprint arXiv:2111.07832}, 2021.

\end{thebibliography}
}

\end{document}